\documentclass[11pt]{article}

\usepackage[preprint]{acl}

\usepackage{times}
\usepackage{latexsym}

\usepackage[T1]{fontenc}

\usepackage[utf8]{inputenc}
\usepackage{amsthm}
\newtheorem{definition}{Definition}

\usepackage{microtype}

\usepackage{inconsolata}

\usepackage{graphicx}

\usepackage[utf8]{inputenc} 
\usepackage[T1]{fontenc}    
\usepackage{hyperref}       
\usepackage{url}            
\usepackage{booktabs}       
\usepackage{amsfonts}       
\usepackage{nicefrac}       
\usepackage{microtype}      
\usepackage{xcolor}         

\usepackage{algorithm}
\usepackage{algorithmic}
\usepackage[utf8]{inputenc} 
\usepackage[T1]{fontenc}    
\usepackage{url}            
\usepackage{booktabs}       
\usepackage{amsfonts}       
\usepackage{nicefrac}       
\usepackage{microtype}      
\usepackage{bm}
\usepackage{enumitem}
\usepackage{graphicx}
\usepackage{amsmath}
\usepackage{subfigure}
\usepackage{amssymb}
\usepackage{caption}
\usepackage{longtable}
\usepackage{soul}
\usepackage{color}
\usepackage{appendix}
\usepackage{comment}
\usepackage{xcolor}         
\usepackage{centernot}

\usepackage{array}
\usepackage{multirow}
\usepackage{wrapfig,lipsum,booktabs}
\usepackage{makecell}
\usepackage{pifont}
\newcommand{\cmark}{\ding{51}}%
\usepackage{tabularx}
\DeclareMathOperator*{\argmin}{arg\,min}
\DeclareMathOperator*{\argmax}{arg\,max}

\newcommand{\hllime}[1]{{\sethlcolor{lime!30}\hl{#1}}}
\newcommand{\hlgreen}[1]{{\sethlcolor{teal!30}\hl{#1}}}
\newcommand{\hlblue}[1]{{\sethlcolor{cyan!30}\hl{#1}}}
\newcommand{\hlmeg}[1]{{\sethlcolor{magenta!30}\hl{#1}}}
\usepackage{newfloat}
\usepackage{listings}

\usepackage{color, colortbl}
\definecolor{Gray}{gray}{0.9}

\newcommand{\titleName}{\texttt{SEE}}
\newcommand{\titleNamewithspace}{\texttt{SEE} }
\newcommand{\frameName}{\texttt{SEE} }
\newcommand{\frameNamenospace}{\texttt{SEE}}
\newcommand{\inputOutputPair}{\textit{input/output pair }}
\newcommand{\exampleInstruction}{\textit{prompt example }}

\newcommand{\exampleInstructionnospace}{\textit{prompt example}}

\newcommand{\phasezero}{\colorbox{cyan!15}{\texttt{//Phase 0}}}
\newcommand{\phaseone}{\colorbox{green!15}{\texttt{//Phase 1}}}
\newcommand{\phasetwo}{\colorbox{orange!15}{\texttt{//Phase 2}}}
\newcommand{\phasethree}{\colorbox{yellow!15}{\texttt{//Phase 3}}}

\title{SEE: Strategic Exploration and Exploitation for Cohesive In-Context Prompt Optimization}

\author{Wendi Cui$^1$\thanks{For correspondence regardng this paper please reach out to yduwcui@gmail.com, or jxzhangai@gmail.com.}, Zhuohang Li$^3$,  Hao Sun $^4$, Damien Lopez$^1$, Kamalika Das$^{1,2}$, \\ {\bf Bradley Malin$^{3,5}$, Sricharan Kumar$^{1,2}$, Jiaxin Zhang$^{1,2*}$}
\\ $^1$Intuit \quad  $^2$Intuit AI Research \quad $^3$Vanderbilt University \quad $^4$ University of Cambridge \\ \quad $^5$Vanderbilt University Medical Center\\
}

\begin{document}
\maketitle

\begin{abstract}
Designing optimal prompts for Large Language Models (LLMs) is a complicated and resource-intensive task, often requiring substantial human expertise and effort. Existing approaches typically separate the optimization of prompt instructions and in-context learning examples, leading to in-cohesive prompts that is defined and represented by suboptimal task performance. To overcome these challenges, we propose a novel Cohesive In-Context Prompt Optimization framework that refines both prompt instructions and examples. However, formulating such an optimization in the discrete and high-dimensional
space of natural language poses significant challenges in both convergence and computational efficiency. To address these issues, we introduce, \frameNamenospace, a scalable and efficient prompt optimization framework that adopts metaheuristic optimization principles and strategically balances exploration and exploitation to enhance optimization performance and achieve efficient convergence. \frameName features a quad-phased design that alternates between global traversal (exploration) and local optimization (exploitation) and adaptively chooses LLM operators during the optimization process. We have conducted a comprehensive evaluation across 35 benchmark tasks, and \frameName significantly outperforms state-of-the-art baseline methods by a large margin, achieving an average performance gain of {\bf 13.94} while reducing computational costs by {\bf 58.67}\%.
\end{abstract}

\vspace{-1mm}
\begin{figure}[ht]
    \centering
    \includegraphics[width=0.9\linewidth]
    {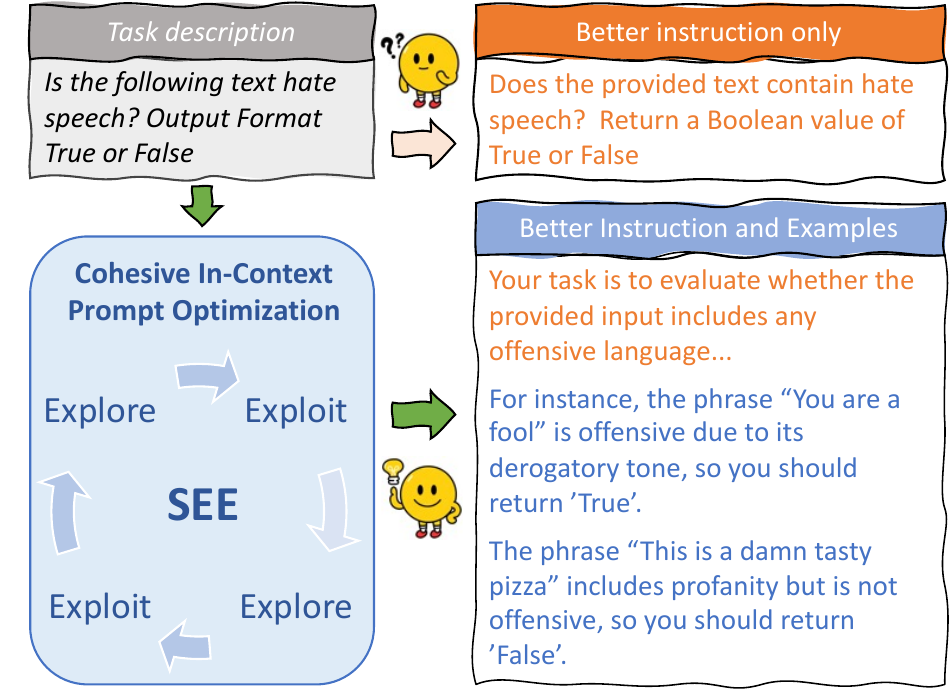}
    \vspace{-2mm}
    \caption{An illustrative example of the cohesive \\ in-context prompt optimization.}
    \label{fig:overview}
\vspace{-3mm}
\end{figure}

\section{Introduction}
Large Language Models (LLMs) have exhibited extraordinary performance across various domains and tasks \cite{bubeck2023sparks,yang2023harnessing}.
Prompt engineering seeks to craft effective prompts that unleash the complete capabilities of LLMs. It is becoming an increasingly popular option for quickly adapting LLMs for downstream tasks due to its compatibility with black-box APIs (e.g., GPT-4 \cite{OpenAI2023GPT4TR} and PaLM \cite{chowdhery2022palm}), and its cost-effectiveness compared to the conventional fine-tuning paradigm. Although good prompt design can substantially improve LLM's performance~\cite{zhu2023promptbench}, manually optimizing prompts is a long process that often requires significant human effort and expert knowledge. Thus automatic prompt optimization is critical to leveraging the power of LLMs and reducing human effort.

However, automatic prompt optimization is a non-trivial task that involves discrete variables and complex high-dimensional spaces \cite{Zhou2023APE}. To avoid optimizing discrete long prompts, existing research treats the optimization of instruction and examples as separate tasks: one line of research~\cite{pryzant2023APO, Chen2023InstuructZero, Yang2023OPRO, Guo2023EVOPrompt} takes the zero-shot prompting approach \cite{kojima2022large} to focus on \textit{optimizing a short instruction} that comprises one or few sentences; while the other line of work~\cite{liu2021makes, lu2021fantastically, lu2022dynamic,zhang2022active,an2023skill} emphasizes more the importance of few-shot examples \cite{Brown2020fewshot-icl} and seeks to \textit{selecting the best set of examples} from a pre-defined dataset given a \textit{fixed} instruction. Although such treatment effectively reduces the optimization complexity, it overlooks the cohesiveness of the full prompt and the interplay between instruction and examples, resulting in {\em sub-optimal} performance \cite{Hsieh2023AELP}.

In this work, we formulate the problem to be a cohesive optimization of instruction and examples where we {\bf simultaneously optimizes the prompt instruction and examples as a whole}. Coherence is defined as the degree to which the prompt components (instruction and examples) work effectively together to achieve strong task results. As illustrated in Figure~\ref{fig:overview}, our goal is to not impose any restrictions or assumptions on the prompt style (zero-shot or few-shot), thereby unlocking the full potential of prompt traversal in contrast to previous instruction-only optimization methods~\cite{Zhou2023APE,pryzant2023APO,Chen2023InstuructZero,Guo2023EVOPrompt,Fernando2023PromptBreeder}. Such a formulation will derive highly adaptive and flexible prompts, ranging from a simple zero-shot prompt to an elaborative few shot prompt with detailed COT examples, depending on the task at hand.

However, such a problem formulation results in a complex combinatorial optimization problem that naturally brings two \textit{challenges}:
(1) \textbf{performance-wise}, how to design an optimization framework that navigates the high-dimensional joint space of instructions and examples, steering clear of local minima to ensure continuous performance enhancement? (2) \textbf{cost-wise}, what strategies can be employed to improve the efficiency of the algorithm, enabling fast convergence with a reasonable level of computational complexity?

To address such challenges, we propose a novel Strategic Exploration and Exploitation (\titleName) framework that aims at accelerating cohesive prompt optimization in high-dimensional spaces while minimizing inference costs. Targeting at continuous \textit{performance enhancement}, \titleNamewithspace adopts the principles of \textbf{metaheuristic optimization framework} which is an iterative refinement framework widely used for complex high dimensional optimizations \cite{talbi2009metaheuristics}. To \textit{reduce the cost}, \frameName introduces a quad-phased design that {\bf strategically alternates between exploration and exploitation}, efficiently navigating high-dimensional space. 

To apply \frameName to prompt optimization task, we identify five LLM operators to generate new candidates in each iteration. By analyzing operators' unique strengths and features,  \frameName is able to \textbf{adaptively choose the best operators} during the optimization process, achieving optimal performance while accelerating convergence speed. Additionally, we integrate two innovative designs to enhance the performance and efficiency of \titleName. Firstly, we introduce a \textit{task-aware similarity metric} using \textit{performance-based vectors} and \textit{hamming distance}, proving more effective than traditional lexical similarity metrics. Secondly, we implement \textit{adaptive phase stop criteria} that ensure maximum performance improvement while optimizing the overall efficiency.

We conduct an extensive evaluation on a total number of $35$ benchmark tasks and empirically show that \frameName demonstrates substantial improvements compared to $9$ state-of-the-art (SOTA) methods, including \textit{APE}~\cite{Zhou2023APE}, \textit{APO} ~\cite{pryzant2023APO}, \textit{OPRO}~\cite{Yang2023OPRO},     \textit{PromptBreeder}~\cite{Fernando2023PromptBreeder}, \textit{EvoPrompt}~\cite{Guo2023EVOPrompt}, \textit{MoP}~\cite{wang2024mop}, \textit{EASE}~\cite{wu2024ease}, \textit{ZOPO}~\cite{huw2024zopo}, and \textit{AELP} ~\cite{Hsieh2023AELP}, with the significant computational cost reduction. For harder tasks like BBH, \frameName introduces an average of \textbf{13.94} task accuracy improvement while reducing \textbf{58.67\%} of computational costs compared to SOTA methods. In summary, our key contributions are:
\begin{itemize}[leftmargin=10pt, itemsep=0.5pt]
    \item We propose \textit{\frameNamenospace}, a novel framework integrating metaheuristic optimization principles to simultaneously optimize instructions and examples as a unity, allowing it to generate both zero-shot and few-shot prompts. To the best of our knowledge 
    \frameName is the first framework with such capability.
    \item We introduce an innovative quad-phase design that strategically balances exploration and exploitation. Together with an adaptive operator selection mechanism that uses the most suitable operator at the right time, such innovation significantly enhances the efficiency compared to traditional metaheuristic optimization frameworks.
    \item We conduct extensive evaluations, demonstrating that \textit{\frameName}achieves substantial improvements over state-of-the-art (SOTA) methods while significantly reducing computational costs.
\end{itemize}

\vspace{-2mm}
\section{Preliminaries}
\vspace{-1mm}

\paragraph{Problem Formulation}
Considering the task $\mathcal{T}$ specified by a dataset $\mathcal{D} = {(\mathcal{Q}, \mathcal{A})}$ of input/output pairs, the LLM $\mathcal{L}$ produces the corresponding output $\mathcal{A}$ via prompting with the concatenation of prompt $\mathcal{P}$ and a given input $\mathcal{Q}$, i.e., $[\mathcal{P}; \mathcal{Q}]$. The objective of prompt optimization is to design the best natural language prompt $\mathcal{P}^*$ that maximizes the performance of $\mathcal{L}$ on $\mathcal{T}$. 

Typically, an ideal prompt $\mathcal{P}$ consists of \emph{instruction}, denoted by $\mathcal{I}$ and \emph{examples} denoted by $\mathcal{E}$ as in-context learning (ICL) demonstrations. Our goal of joint prompt optimization is to search for the optimal prompt $\mathcal{P}_{(\mathcal{I},\mathcal{E})}^*$ given $\mathcal{L}$ that maximizes the performance towards a performance metric function $\mathcal{F}$ (e.g., accuracy). This can be formally defined as the following optimization problem:
\begin{equation}
    \mathcal{P}_{(\mathcal{I},\mathcal{E})}^* = \argmax_{\mathcal{P}_{(\mathcal{I},\mathcal{E})} \in \mathcal{X}} \mathbb{E}_{(\mathcal{Q}, \mathcal{A})} \left[ {\mathcal{F}}(\mathcal{P}_{(\mathcal{I},\mathcal{E})}; \mathcal{Q}, \mathcal{A}) ~|~ \mathcal{L} \right], \label{eq:formualtion}
\end{equation}

where $\mathcal{X}$ denotes the sample space for a natural language prompt, a discrete and intractable space of arbitrarily large dimension, which makes the optimization problem in Eq. \eqref{eq:formualtion} extremely difficult.

\paragraph{Metaheuristic Optimization Framework}

The \textit{metaheuristic optimization framework} provides a generalized approach for solving complex optimization problems, particularly those involving high-dimensional or non-convex solution spaces where traditional methods may struggle \cite{talbi2009metaheuristics}. The framework typically follows an iterative process comprising the following key components:

\begin{itemize}[leftmargin=10pt, itemsep=0.1pt]
    \item \textbf{Initialization:} An initial set of candidate solutions is generated, often randomly or using heuristic methods to ensure a diverse starting candidate pool.
    \item \textbf{Generation and Variation:} New candidate solutions are derived through \textit{Operators} such as mutation, crossover, probabilistic sampling, or local search, facilitating effective exploration of the solution space.
    \item \textbf{Selection and Pruning:} Candidates are evaluated using an objective function, and suboptimal solutions are discarded to refine the search toward optimal or near-optimal results.
\end{itemize}

This iterative process continues until a termination criterion, such as convergence to a solution or reaching a computational limit, is met. Examples of metaheuristic methods include \textit{Genetic Algorithms (GA)},
which simulate the process of natural evolutio; 
\textit{Particle Swarm Optimization (PSO)}, inspired by the social behavior of birds or fish to iteratively refine solutions; and \textit{Differential Evolution (DE)}, which optimizes by iteratively combining and mutating candidate solutions. 
These techniques are widely applied in fields such as engineering design, scheduling, and machine learning \cite{talbi2009metaheuristics, blum2003metaheuristics}.

\section{Proposed Method: Strategic Exploration and Exploitation (\frameNamenospace) }

\subsection{Intuition}
The intuition behind our proposed framework, \frameNamenospace, lies in addressing key limitations of traditional metaheuristic algorithms. Existing methods often apply generation and variation in a repetitive and uniform manner—such as genetic algorithms relying on mutation and crossover repeatedly—without adapting to the specific needs of the optimization process. This introduces unnecessary randomness, increasing computational costs and slowing convergence. In contrast, our framework strategically divides the optimization process into \textit{four distinct phases}, each dedicated to either exploration or exploitation, thereby accelerating the overall process.

To maximize efficiency in each phase, we ensure that only the most effective LLM operators tailored to the requirements of each phase, are utilized to generate new candidates.  By strategically \textit{applying the right operator at the right time}, \frameName achieves both faster convergence and improved performance, delivering a cohesive combination of instructions and examples for a variety of tasks.

\begin{figure}[t]
    \centering
    \includegraphics[width=0.82\linewidth]
    {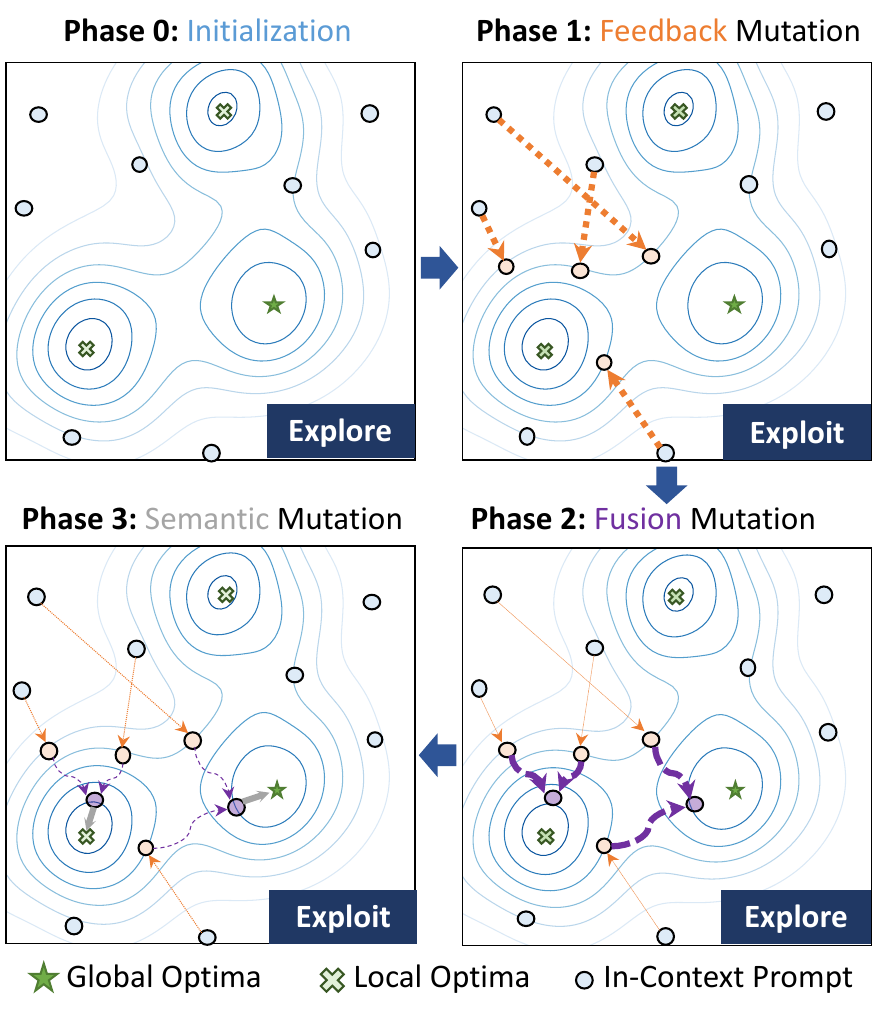}
    \vspace{-0.3cm}

    \caption{Illustration of \frameName framework.}
    \label{fig:localize}
\end{figure}
\vspace{-1mm}

\subsection{Generation Operators}
\frameName uses LLM operators to perform generation and variation. Full prompts of the operators can be found in the Appendix section \ref{sec:operator-prompts}. For operators designed for exploration, we call them global operators. For ones best at exploitation, we define them as local operators. \frameName utilizes five operators that are used in prior research. More detailed operator information can be found in Appendix section \ref{sec:operator-def}. The three {\em global} operators are:

\begin{itemize}[leftmargin=10pt, itemsep=0.5pt, topsep=1.5pt, partopsep=1.5pt]
    \item {\bf Lamarckian Operator} is a reverse-engineering operator $\mathcal{O}_{L}$ that accepts \textit{input-output} pairs of the task and attempts to ``reverse-engineer'' the task instruction which will generate the \textit{output} based on the \textit{input}.  
    \item {\bf Estimation of Distribution Operator (EDA)} is a function operator $\mathcal{O}_{E}$ that takes in a group of candidates and outputs a new candidate by studying the input group. If the input group is ranked from the best performer to the worst performer, we call it EDA + Index (EDA+I).   
    \item {\bf Crossover Operator (CR)} is a function operator $\mathcal{O}_{C}$ that takes two candidates as parents and generates a new candidate mixing the traits of both parents. If the parents are chosen by prioritizing differences between them, we call it Crossover + Distinct (CR + D).
\end{itemize}

The two {\em local} operators are:

\begin{itemize}[leftmargin=10pt, itemsep=0.5pt, topsep=1pt]
    \item {\bf Feedback Operator} is a function operator $\mathcal{O}_{F}$ that utilizes two LLM agents. $\mathcal{O}_{F}$ first passes mistakes an existing candidate makes to an ``Examiner'' agent, whose task is to examine these mistakes and provide remediation strategies. It then uses an ``Improver'' agent that takes the remediation strategies and applies them to the existing candidate to generate a new candidate.
    \item {\bf Semantic Operator} is a function operator $\mathcal{O}_{S}$ that takes in an existing candidate, and modifies the candidate lexically while preserving its semantic meaning.
\end{itemize}

To better harness these operators, we compare them along the following five dimensions that are critical to our exploration-exploitation strategy in terms of performance and efficiency:

\begin{itemize}[leftmargin=10pt, itemsep=0.5pt, topsep=1.5pt, partopsep=1.5pt]
    \item \textbf{Add or remove examples.} This examines whether an operator can add or remove few-shot examples, to traverse the entire space of a cohesive prompt optimization problem. 
    
     \item \textbf{Probability of improvement.} This evaluates the probability (successful rate) of an operator that brings performance improvement after iterations. 

    \item \textbf{Convergence speed.} This metric aims to evaluate how fast (in terms of iterations) an operator needs to optimize the current candidate to its local minimum solution. 

    \item \textbf{Two or more parents?} This indicates whether an operator needs two or more input candidates(parents) to generate a new candidate. Operators needing more than one parent have the potential to combine traits from diverse parents, enhancing global exploration capability. 

    \item \textbf{API cost per operation.} It is the number of API calls needed to perform a specific operator via LLM agents. 
\end{itemize}

We conducted a series of experiments (ran each operator 100 times based on 4 different initialization settings) to assess the performance of each operator regarding the five features, aiming at obtaining a comprehensive understanding of the inherent strengths and weaknesses of each operator. This allows us to select effective operators to find optimal solutions in an accelerated manner. As shown in Table \ref{tab:operator}, we observe that the Lamarckian operator is a crucial operator that introduces diverse samples, making it an ideal choice for exploration and global initialization. The feedback operator leads to faster convergence (four $\bullet$), making it good for rapid exploitation, but it requires two API/inference calls (two $\bullet$), higher than the other operators (one $\bullet$). EDA and Crossover operators share similar characteristics that combine traits from multiple parents and lead to a higher probability of improvement (four $\bullet$), indicating their excellence in exploring the global space. For a more in-depth discussion on operators, please refer to Appendix \ref{sec:op_fewshow} and \ref{sec:op-feature}.

\begin{table}[t]
\vspace{-0.3cm}
\centering
\resizebox{\linewidth}{!}{
\begin{tabular}{@{}c|ccc|ccc@{}}
\toprule
Operator           & Add & Remove & Parents & Prob & Speed  & Cost\\ 
\midrule
Lamarckian & \cmark    & -  &  - &  - & -  & $\bullet$ \\
Feedback   & \cmark    & \cmark  & - & $\bullet~\bullet$  & $\bullet\bullet\bullet\bullet$ & $\bullet\bullet$\\
EDA   & -      & -  & \cmark & $\bullet\bullet\bullet \bullet $  & $\bullet\bullet$  & $\bullet$\\
Crossover         & -      & -  & \cmark & $\bullet\bullet\bullet \bullet $  & $\bullet\bullet$  & $\bullet$\\
Semantic   & -     & \cmark  & - & $\bullet\bullet\bullet$   & $\bullet\bullet\bullet$ & $\bullet$\\ 
\bottomrule
\end{tabular}
}
\caption{Qualitative analysis of mutation operators}
\label{tab:operator}
\end{table}

\begin{algorithm*}[h!]
\small
\begin{algorithmic}[1]
\STATE {\bf requirements}: size of pool $n$, a dev set $\mathcal{D}_{\textup{dev}}$, score function $\mathcal{F}$ on the base LLM $\mathcal{L}$, phase improvement $t$ and performance gain threshold $t^*$ and minimum run time tolerance for phases $\mathcal{K}_i$, designed operators $\mathcal{O}_{L}$, $\mathcal{O}_{F}$, $\mathcal{O}_{E}$, $\mathcal{O}_{C}$ and $\mathcal{O}_{S}$
\STATE 
\colorbox{cyan!15}{
\parbox{\dimexpr\linewidth-4\fboxsep}{
{\bf initialization}: generate diverse initial prompts 
$\mathcal{P}^0 = \{p_1^0, ..., p_n^0\}$ by $\mathcal{O}_{l}$ or $\mathcal{O}_{s}$, evaluate initial scores $\mathcal{S}^0 \leftarrow \{s_i^0 = \mathcal{F}(p_i^0, \mathcal{D}_{\textup{dev}})\}$
\hfill {\phasezero}
}
›}

\STATE 
\colorbox{green!15}{
\parbox{\dimexpr\linewidth-4\fboxsep}{
{\bf while} $t < t^*$ or $k \le \mathcal{K}_1$ {\bf do}
\STATE \quad {\bf \em Local Feedback Operation:} generate new prompts 
by Feedback Operator, $\mathcal{P}_t \leftarrow \mathcal{O}_{f}(\mathcal{P}^0)$, 
evaluate $\mathcal{S}_t \leftarrow \mathcal{F}(\mathcal{P}^0, \mathcal{D}_{\textup{dev}})$,
update $\mathcal{P}^1 \leftarrow \{\mathcal{P}_t, \mathcal{P}^0 \}$, 
and score set $\mathcal{S}^1 \leftarrow \{\mathcal{S}_t, \mathcal{S}^0 \}$
\hfill {\phaseone}
}
}

\STATE 
\colorbox{orange!15}{
\parbox{\dimexpr\linewidth-4\fboxsep}{
{\bf while} $t < t^*$ or $k \le \mathcal{K}_2$ {\bf do}
\STATE \quad {\bf \em Global Fusion Operation:} select prompts from the current pool 
$\{p_{r_1},...,p_{r_k}\} \in \mathcal{P}^1$, generate a new prompt via EDA or Crossover Operators, 
evaluate $s_t \leftarrow \mathcal{F}(p_t, \mathcal{D}_{\textup{dev}})$, and update 
$\mathcal{P}^2 \leftarrow \{\mathcal{P}^1, p_t\}$ and $\mathcal{S}^2 \leftarrow \{\mathcal{S}^1, s_t\}$ 
\hfill {\phasetwo}
}}

\STATE 
\colorbox{yellow!15}{
\parbox{\dimexpr\linewidth-4\fboxsep}{
{\bf while} $t < t^*$ or $k \le \mathcal{K}_3$ {\bf do}
\STATE \quad {\bf \em Local Semantic Operation:} generate new prompts by Semantic Operator $\mathcal{P}_t^* \leftarrow \mathcal{O}_{s}(\mathcal{P}^2)$, 
evaluate $\mathcal{S}_t^* \leftarrow \mathcal{F}(\mathcal{P}^2, \mathcal{D}_{\textup{dev}})$, 
and update $\mathcal{P}^3 \leftarrow \{\mathcal{P}_t^*, \mathcal{P}^2 \}$, 
and $\mathcal{S}^3 \leftarrow \{\mathcal{S}_t^*, \mathcal{S}^2 \}$
\hfill {\phasethree}
}}

\STATE {\bf return} $p^* \leftarrow \argmax_{p \in \mathcal{P}^3} \mathcal{F}(p, \mathcal{D}_{\textup{dev}})$
\end{algorithmic}
\caption{\frameName Framework}
\label{algo_self}
\end{algorithm*}

\subsection{\frameName Framework} 
The \frameName framework approaches the complex optimization problem strategically through four distinct phases. Beyond the operators mentioned above, it requests up to three data sets. $\mathcal{D}_{\textup{train}}$ is used for the first phase of  initialization, specifically used by the Lamarckian Operator. $\mathcal{D}_{\textup{dev}}$ acts as a development data set to compute the performance score for each candidate during the optimization process. $\mathcal{D}_{\textup{test}}$ is used for the final performance evaluation of the optimized prompt.

\subsubsection{Phase 0: Global Initialization}

Following the principle of metaheuristic optimization, phase 0 aims to create diverse candidates as the initial candidate pool to explore the vast joint space of instruction and example. We provide two types of initialization based on the availability of data: initializing from \inputOutputPair of the task, denoted \frameNamenospace-io-pair, and initializing from human-composed example prompts, denoted \frameNamenospace-example. 

\begin{itemize}[leftmargin=10pt, itemsep=0.5pt, topsep=1.5pt, partopsep=1.5pt]
    \item {\bf \frameNamenospace-io-pair:} Given a set of input/output pairs $S = \{(Q_1, A_1),...,(Q_m,A_m)\}$ from $\mathcal{D}_{\textup{train}}$, representing the input and output for the task $\mathcal{T}$, \frameName apply  Lamarckian Operator $\mathcal{O}_{L}$ to {\em reverse engineer} potential prompts from provided demonstrating pairs. 
    
    \item {\bf \frameNamenospace-example:} \frameName takes expert constructed prompts and apply Semantic Operator $\mathcal{O}_S$ to enhance the diversity of the initial candidate pool. This allows humans to jump-start the optimization process by incorporating prior knowledge.

\end{itemize}

\subsubsection{Phase 1: Local Feedback Operation}
\vspace{-1mm}

Phase 1 to Phase 3 adheres to the metaheuristic optimization principles where each phase first conducts generation and variation through designated operators, then performs selection and pruning greedily based on the candidates' performance score on the development set $\mathcal{D}_{\textup{dev}}$. 

While an initial phase (Phase 0) may result in a diverse candidate pool, each candidate could still be distant from the best version of itself, its local minimum. To address this, \frameName exploits each candidate by employing the Feedback Operator $\mathcal{O}_{F}$ to expedite its convergence towards their local minimums. This involves the introduction of an LLM {\em Examiner} to generate bespoke improvement guidance and an LLM {\em Improver} to apply these to generate new candidates.

\vspace{-1mm}
\subsubsection{Phase 2: Global Fusion Operation}
\vspace{-1mm}

Phase 1 provides a more refined set of candidates, while some of them might be stuck in local optima. To address this issue, we prioritize exploration rather than exploitation in Phase 2. By performing fusion between different candidates leveraging EDA (EDA-I) Operators $\mathcal{O}_E$ and CR (CR-D) Operators $\mathcal{O}_C$ which request multiple parents,  \frameName facilitates the increased fusion of traits among candidates on a larger global scale, thus enabling escape from these local optima. Rather than employing cosine similarity as distance metrics, we adopt \textit{hamming distance} (see more discussions in Section \ref{sec-sim}) for calculating similarity on performance-based vectors so that more diversity is promoted during optimization.  

\vspace{-1mm}
\subsubsection{Phase 3: Local Semantic Operation}
\vspace{-1mm}

Upon completing Phase 2's exploration, Phase 3 employs local exploitation to hasten the ``last mile'' of convergence. As the concluding phase of \frameNamenospace, the performance score of the candidate pool is relatively optimized. The Semantic Operator $\mathcal{O}_S$ is selected to expedite a more cost-effective exploitation. Finally, we identify the best candidate as our ultimate optimal prompt and assess its performance on the testing dataset $\mathcal{D}_{\textup{test}}$. The workflow of \frameName framework is shown in Algorithm \ref{algo_self}. 

\vspace{-2mm}
\subsection{\frameName Novel Design Schemes}
\label{sec-sim}
We also propose two novel design schemes to improve performance and efficiency.  

\begin{table*}[h!]
\vspace{-0.3cm}
\resizebox{\linewidth}{!}{
\small
\begin{tabular}{@{}l|llllllll@{}}
\toprule
Method       
& \begin{tabular}[l]{@{}l@{}}Causal\\ Judgement\end{tabular} 
& \begin{tabular}[l]{@{}l@{}}Dis\\-ambiguation\end{tabular}  
& \begin{tabular}[l]{@{}l@{}}Dyck \\ Languages\end{tabular} 
& \begin{tabular}[l]{@{}l@{}}Formal \\Fallacies\end{tabular} 
& Hyperbaton 
& \begin{tabular}[l]{@{}l@{}}Logical\\ Five\end{tabular} 
& \begin{tabular}[l]{@{}l@{}}Color\\ Reasoning\end{tabular} 
& \begin{tabular}[l]{@{}l@{}}Salient \\ Translation\end{tabular} 
\\ 
\midrule

OPRO \cite{Yang2023OPRO}        
& 71.94
& 71.53
& 36.73
& 49.51
& 75.92
& 50.00
& 65.55
& 43.88
\\ 

EvoPrompt \cite{Guo2023EVOPrompt} 
& 67.24
& 53.70
& \bfseries 47.96
& 50.81
& 74.79
& 61.40
& 60.90
& 47.58
\\ 

AELP \cite{Hsieh2023AELP}        
& 77.77
& 64.79
& 10.67
& 58.25
& 53.74
& 73.49
& 68.14
& 41.43
\\ 
\midrule

\frameNamenospace-io-pair  
& 72.13
& \bfseries 72.37
& 8.06
& \bfseries 58.87
& 86.02
& 48.19
& 60.52
& \bfseries 49.19
\\

\frameNamenospace-example 
& \bfseries 89.09
& 68.47
& 46.77
& 58.65
& \bfseries 87.50
& \bfseries 86.29
& \bfseries 80.64
& 47.59
\\

\bottomrule
\end{tabular}
}
\caption{Testing performance of the optimal prompt on 8 representative tasks from BBH.}
\label{tab:aelp_datasets}
\end{table*}

\begin{figure*}[h!]
\centering
    \includegraphics[width=0.32\textwidth]{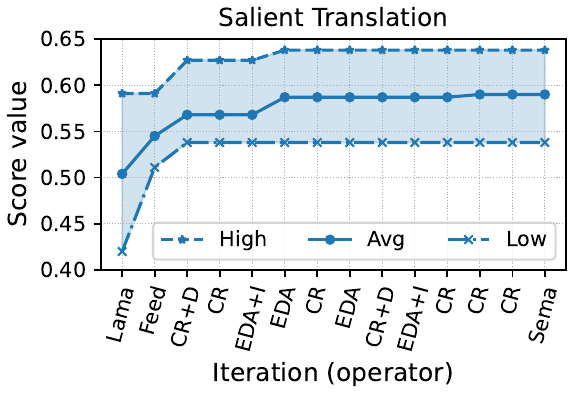}
    \includegraphics[width=0.32\textwidth]{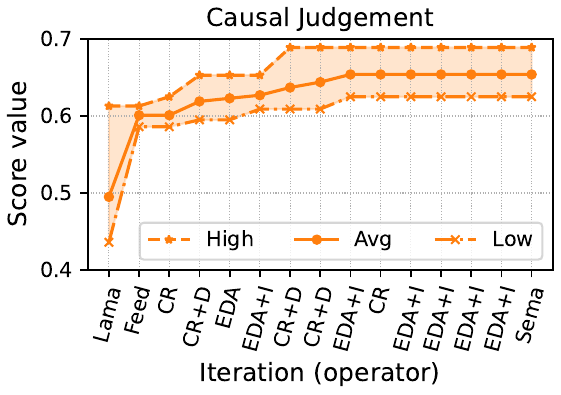}
    \includegraphics[width=0.32\textwidth]{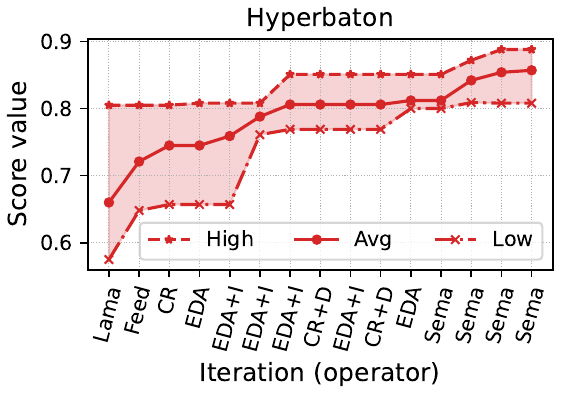}
    \vspace{-2mm}
    \caption{Iteration history of score values with different operators during optimization. The score is defined as the accuracy for the given task on the development set $\mathcal{D}_{\textup{dev}}$.}
    \label{fig:score}
\end{figure*}

\noindent \underline{\textit{1. Performance vector with Hamming distance.}}  
Fusion operators like EDA and Crossover function optimally when parents exhibit distinct attributes \cite{Fernando2023PromptBreeder}. In terms of evaluating similarity scores, we adhere to the principle that similarity should be gauged based on the \emph{performance of the prompts} rather than their linguistic or semantic similarities.  Inspired by this intuition, we propose to construct candidate vectors based on their performance on the development set $\mathcal{D}_{\textup{dev}}$, named ``performance vectors''. To exemplify, in a development dataset comprising five elements, a candidate answering the first three queries correctly and the final two incorrectly would feature a vector representation of $[1, 1, 1, 0, 0]$. 

Rather than calculating the cosine similarity of embedding space, we propose to compute candidate similarity scores by {\em Hamming distance}, which calculates the distance between two vectors of equal length by examining the number of positions at which the corresponding symbols are different. This way ensures that one candidate is more likely to be paired with a candidate that \emph{does not repeat the same mistakes}. 

\label{sec:stop-criteria}

\noindent \underline{\textit{2. Adaptive Phase Stop Criteria.}} To ensure that each optimization phase is fully conducted before transitioning to the next, the decision to proceed to the following phase is influenced by two criteria.

\begin{itemize}[leftmargin=10pt, itemsep=0.1pt, topsep=1.5pt, partopsep=1.5pt]
\item \textit{Performance Gain.} If no performance gain manifests after applying the operators in a particular phase, it's indicative that the candidates have been thoroughly optimized by the operator. Consequently, we transition to the next phase. 

\item \textit{Operator-specific Tolerance.} Not all operators are created equal. For local operators with high convergence speed like the Feedback Operator, it makes sense to transition to the next phase without performance improvements. However, global operators, e.g., Fusion Operators, might not bring immediate improvement but are capable of accessing more diverse parents with traits worth exploring. Therefore, we assign greater \textit{tolerance} to global operators, allowing them to run for a predefined duration even if immediate improvement is not observed. More details about the stop criteria can be found in Appendix \ref{sec:op-feature}.
\end{itemize}

\section{Experiments}
We evaluate \frameName on 35 tasks across 9 baselines. Unless specified, all \frameName results are from GPT-3.5-turbo. 
For additional information regarding the experiment setup, please refer to Appendix \ref{sec:details_exp}.

\paragraph{35 Tasks and Datasets}  We curate 35 benchmark tasks from three domains for thorough experiments: 8 Big Bench Hard (BBH) \cite{suzgun2022challenging}; 3 NLP detection tasks, including Ethos \cite{Mollas2021Ethos}, Liar \cite{Wang2017Liar}, and Sarcasm \cite{Farha2020Sar}; 24 instruction induction tasks \cite{Honovich2022Instuct}. The task and dataset details are in Appendix \ref{sec:tasks}.

\paragraph{9 Baselines} We evaluate \frameName against a variety of LLM-based approaches that have achieved state-of-the-art performance in prompt optimization:   

\begin{itemize}[leftmargin=10pt, itemsep=0.5pt, topsep=1.5pt, partopsep=1.5pt]
    \item {\bf APE} \cite{Zhou2023APE}, {\bf ZOPO} \cite{huw2024zopo} and {\bf APO} \cite{pryzant2023APO}: APE utilizes a Monte Carlo Search strategy that emphasizes {\em exploration}, while APO emphasizes {\em exploitation}, which harnesses incorrect instances as feedback gradient. ZOPO utilizes zeroth-order optimization methods to find local optimal.
    \item {\bf OPRO} \cite{Yang2023OPRO}: OPRO leverages LLM as optimizers to generate better instruction via meta-prompt, solution-score pairs, and task descriptions. 
    \item {\bf PromptBreeder} \cite{Fernando2023PromptBreeder}, {\bf EvoPrompt} \cite{Guo2023EVOPrompt} and {\bf AELP} \cite{Hsieh2023AELP}: these methods connect LLMs with evolution algorithms for prompt optimization. 
    \item {\bf MoP} \cite{wang2024mop}, {\bf EASE} \cite{wu2024ease}: these methods can optimize instructions and examples simultaneously.

\end{itemize}

\subsection{Main Results}
\paragraph{BBH Tasks} 
Following the practice of AELP \cite{Hsieh2023AELP}, we conduct 8 BBH tasks to evaluate the performance of \frameName holistically. We consider two initialization schemes \frameNamenospace-io-pair and \frameNamenospace-example and report the final results in Table \ref{tab:aelp_datasets}. \frameName demonstrates substantial improvements compared to state-of-the-art methods, introducing big average performance increase over AELP (+{\bf 15.31}), EvoPromopt (+{\bf 13.29}), and OPRO (+{\bf 13.21}). 

Fig. \ref{fig:score} depicts the iterative history of prompt optimization, emphasizing the performance score variations for the best, worst candidate, and average candidate performance across iterations. Feedback Operator yields a performance boost within a single iteration and rarely introduces continual improvements. Fusion Operators such as EDA and Crossover aid in escaping local minima and offering additional performance leaps (refer to Hyperbaton). This observation aligns with our initial operator analysis. 
The success of \frameName lies in the organic organization of these operators, harnessing their advantages to maximize performance.

\paragraph{Detection Tasks} 
To present a more expansive comparison, we adopted the configuration outlined in APO \cite{pryzant2023APO} and conducted a comparative analysis against it across three tasks. \frameName exhibits marginally superior performance to APO in relatively simple tasks such as Ethos (+{\bf 1}) and Sarcasm (+{\bf 4}). However, for more complex tasks such as Liar, \frameName demonstrates a significant improvement (+{\bf 18}). Full experiment results are in Table \ref{tab:apo_datasets} in Appendix.

\paragraph{Instruction Induction Tasks} 

To compare \frameName with broader sets of baselines, we evaluate \frameName on APE's 24 instruction induction tasks. The results show that \frameName outperforms in 87.5\% tasks over APE and MoP, 91.7\% tasks over PromptBreeder, 100\% tasks over Evoprompt, OPRO, ZOPO, and 66.7\% tasks over EASE.
Table \ref{tab:apetask} in Appendix \ref{sub:apetask} provides complete experimental results. 

\subsection{Analysis}

\paragraph{Applicability of \frameName framework} 
To evaluate the general applicability of the \frameName framework, we perform end-to-end optimizations on a diverse set of models, covering both open-source and closed-source LLMs. Each model undergoes three end-to-end runs, with the average performance and standard deviation reported. As shown in Table \ref{tab:bbh-models}, GPT-4 consistently achieves the highest performance across all tasks, followed by Llama3-70B. Claude 2 demonstrates comparable performance to GPT-3.5. For open-source LLM models, Mistral-7B and Llama3-8B are comparable to each other, both outperforming Llama2-7B by a large margin.

\begin{table}

\resizebox{\linewidth}{!}{
\begin{tabular}{@{}lllll@{}}
\toprule
Model                                                                                                   & \begin{tabular}[l]{@{}l@{}}Dis- \\      ambiguation\end{tabular}  & \begin{tabular}[l]{@{}l@{}}Formal \\Fallacies\end{tabular} & Hyperbaton       & \begin{tabular}[l]{@{}l@{}}Salient \\      Translation\end{tabular} \\ \midrule
GPT-3.5               & $69.99_{(2.95)}$ & $58.49_{(0.33)}$ & $84.35_{(1.83)}$ & $48.39_{(0.66)}$    \\
GPT-4  & $79.34_{(3.33)}$ & $75.91_{(0.53)}$ & $90.58_{(1.39)}$ & $70.45_{(0.99)}$    \\
PaLM 2               & $71.49_{(0.37)}$ & $58.33_{(1.53)}$ & $79.45_{(0.98)}$ & $49.07_{(3.25)}$    \\
Claude 2                                                                                             & $72.95_{(2.26)}$ & $49.46_{(1.52)}$ & $83.32_{(1.01)}$ & $61.82_{(0.38)}$    \\  \midrule
Mistral-7B          & $65.89_{(0.76)}$ & $53.23_{(1.74)}$ & $78.76_{(1.36)}$ & $43.84_{(1.00)}$    \\
Llama2-7B            & $42.74_{(4.61)}$ & $56.72_{(1.37)}$ & $53.23_{(2.37)}$ & $21.23_{(1.01)}$    \\
Llama3-8B            & $62.63_{(3.85)}$ & $71.50_{(4.85)}$ & $57.52_{(4.28)}$ & $37.09_{(2.86)}$    \\ 
Llama3-70B           & $74.73_{(2.01)}$ & $70.93_{(2.25)}$ & $82.26_{(0.66)}$ & $62.90_{(1.97)}$  \\  
\bottomrule
\end{tabular}
}
\caption{\frameName performance with different LLM models}
\label{tab:bbh-models}
\end{table}

\begin{table}
\resizebox{\linewidth}{!}{
\begin{tabular}{@{}lllll@{}}
\toprule
Method                                                                                                   & \begin{tabular}[l]{@{}l@{}}Dis- \\      ambiguation\end{tabular}  & \begin{tabular}[l]{@{}l@{}}Formal \\Fallacies\end{tabular} & Hyperbaton       & \begin{tabular}[l]{@{}l@{}}Salient \\      Translation\end{tabular} \\ \midrule
OPRO                                                    & 71.53                           & 49.51                           & 75.92                           & 43.88                           \\
OPRO-fs                                             & \cellcolor{orange!25}66.93                           & \cellcolor{cyan!25}52.41                           & \cellcolor{orange!25}62.90                            & \cellcolor{orange!25}37.39                           \\
EvoPrompt                                                                                                                                    & 53.7                            & 50.81                           & 74.79                           & 47.58                           \\
EvoPrompt-fs                                        & \cellcolor{cyan!25}57.43                           & \cellcolor{orange!25}43.54                           & \cellcolor{cyan!25}79.83                           & \cellcolor{orange!25}31.45                           \\
 \midrule
\frameNamenospace-io-pair                   & \textbf{72.37} & \textbf{58.87} & 86.02                           & \textbf{48.19} \\
\frameNamenospace-example & 68.47                           & 58.65                           & \textbf{87.51} & 47.59                           \\
\bottomrule
\end{tabular}
}
\caption{Effect of few-shot (fs) examples on BBH tasks.}
\label{tab:bbh-fewshot}
\end{table}

\paragraph{Necessity of Cohesive Prompt Optimization} 
To better understand whether cohesive prompt optimization is necessary, we randomly add two few-shot examples to OPRO and EvoPrompt. Our results in Table \ref{tab:bbh-fewshot} indicate that OPRO exhibits a performance gain on only 1 / 4 tasks while EvoPrompt shows improvement in 2 / 4 tasks. This suggests the necessity of cohesive prompt optimization as performance degrades if optimized instructions do not align cohesively with naive few-shot selection. 

\begin{table*}[ht]
\centering

\resizebox{\linewidth}{!}{
\begin{tabular}{@{}l|ll|ll|ll|ll@{}}
\toprule
\multirow{2}{*}{Method} & \multicolumn{2}{l|}{Causal Judgement}& \multicolumn{2}{l|}{Disambiguation} & \multicolumn{2}{l|}{Hyperbaton} & \multicolumn{2}{l}{Salient Translation} \\ 
                        & Average score         & High score        & Average score              & High score            & Average score           & High score           & Average score           & High score        \\ \cmidrule(r){1-9}
Random Evo       & $67.70_{(0.75)}$         & $70.28_{(0.56)}$       & $58.22_{(2.47)}$            & $61.3_{(3.17)}$            & $83.00_{(0.15)}$          & $87.8_{(0.00)}$           & $52.00_{(2.35)}$         & $56.80_{(1.60)}$        \\ 
\frameName       & $\textbf{69.88}_{(2.17)}$        & $\textbf{72.00}_{(3.09)}$        & $\textbf{60.32}_{(2.73)}$             & $\textbf{62.9}_{(2.56)}$           & $\textbf{83.52}_{(0.71)}$          & $\textbf{87.8}_{(0.00)}$         & $\textbf{53.06}_{(0.80)}$         & ${\bf 56.80}_{(0.80)}$        \\

\bottomrule
\end{tabular}
}
\caption{Comparison of our phase optimization with traditional random optimization.}
\label{tab:phased-ea}
\end{table*}

\begin{table*}[ht]
\centering

\resizebox{\linewidth}{!}{
\begin{tabular}{@{}l|ll|ll|ll|ll@{}}
\toprule
\multirow{2}{*}{Method} & \multicolumn{2}{l|}{Causal Judgement}& \multicolumn{2}{l|}{Disambiguation} & \multicolumn{2}{l|}{Hyperbaton} & \multicolumn{2}{l}{Salient Translation} \\ 

                        & Average score          & High score        & Average score          & High score        & Average score          & High score           & Average score          & High score     \\ \cmidrule(r){1-9}
Cosine distance       & $64.70_{(2.31)}$         & $67.86_{(2.47)}$       & $58.96_{(1.47)}$            & $63.30_{(0.00)}$            & $74.70_{(1.60)}$          & $85.7_{(0.00)}$           & $49.56_{(1.07)}$         & $58.80_{(0.00)}$        \\ 
Hamming distance        & $\textbf{65.74}_{(2.87)}$        & $\textbf{69.60}_{(2.97)}$        & $\textbf{64.11}_{(1.28)}$             & $\textbf{66.94}_{(2.88)}$           & $\textbf{79.30}_{(4.48)}$          & $\textbf{86.78}_{(2.15)}$         & $\textbf{50.33}_{(2.32)}$         & $\textbf{58.80}_{(0.00)}$        \\
\bottomrule
\end{tabular}
}
\vspace{-2mm}
\caption{Performance comparison of hamming distance and cosine similarity.}
\label{tab:hamming}
\end{table*}

\paragraph{Phase Optimization vs Random Optimization}  
To evaluate the phased design of \frameNamenospace, we compare it against a random optimization strategy on 4 BBH tasks shown in Table \ref{tab:phased-ea}. Notably, \frameName consistently outperforms random optimization in the average score across all tasks and achieves better highest score in two out of four tasks. This superior performance highlights the effectiveness of the well-structured phases with designated operators employed in \frameName.

\begin{figure}[h!]
\centering
    \includegraphics[width=0.8\linewidth]{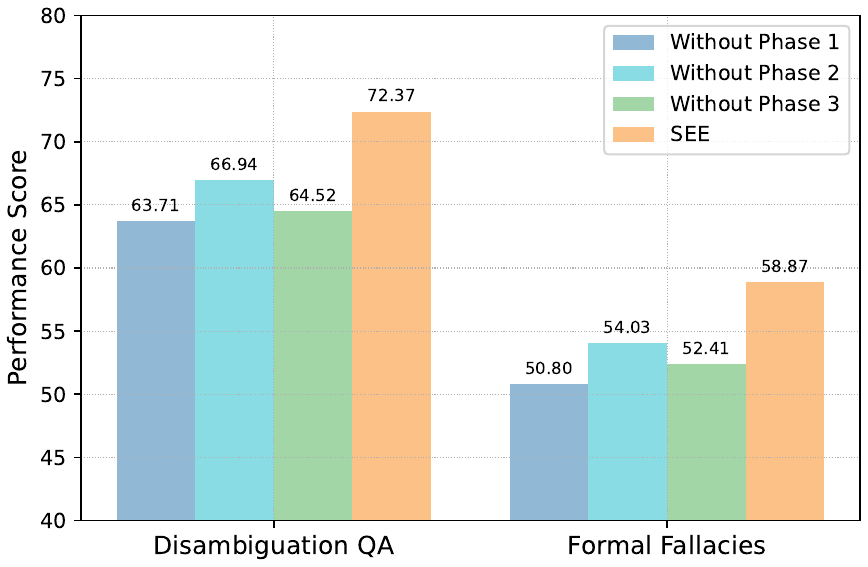}

    \caption{Performance comparison with phases removed}
    \label{fig:phase-performance}
\end{figure}

\paragraph{Effect of Different Phases}
We conducted additional studies to highlight the value of different phases by removing them from the optimization pipeline, as shown in Figure \ref{fig:phase-performance}. We only experiment with Phase 1 -  Phase 3 and did not remove Phase 0 as it generates the initial population. Without Phase 0 there would be no candidate to optimize. We observe no significant differences when different phases are removed. However, removing Phase 1 with the Feedback Operator will cause the greatest performance degradation. We hypothesize that the Feedback Operator allows candidates to arrive at their local optimal efficiently. Thus, removing it will cause the next phase to start with less than locally optimized candidates, impacting the overall performance most. Having all phases yield the best results. 
This further proves the effectiveness and cohesion of the different phases of \frameNamenospace.

\paragraph{Effect of Hamming Distance} We investigate the effectiveness of Hamming distance on performance-based vectors in comparison to the traditional cosine distance applied to embedding vectors for similarity measurement. This analysis is conducted across four optimization iterations. Table \ref{tab:hamming} summarizes the results from four BBH tasks. The findings show that performance vectors using Hamming distance consistently outperform embedding-based approaches using cosine similarity, achieving higher average and maximum scores—particularly in tasks such as Disambiguation (+\textbf{5.2}) and Hyperbaton (+\textbf{4.6}). These results validate the effectiveness of performance-based representations with Hamming distance in improving search efficiency and enhancing task performance.

\begin{figure}[h!]
\centering
\vspace{3.2mm}
    \includegraphics[width=0.49\textwidth]{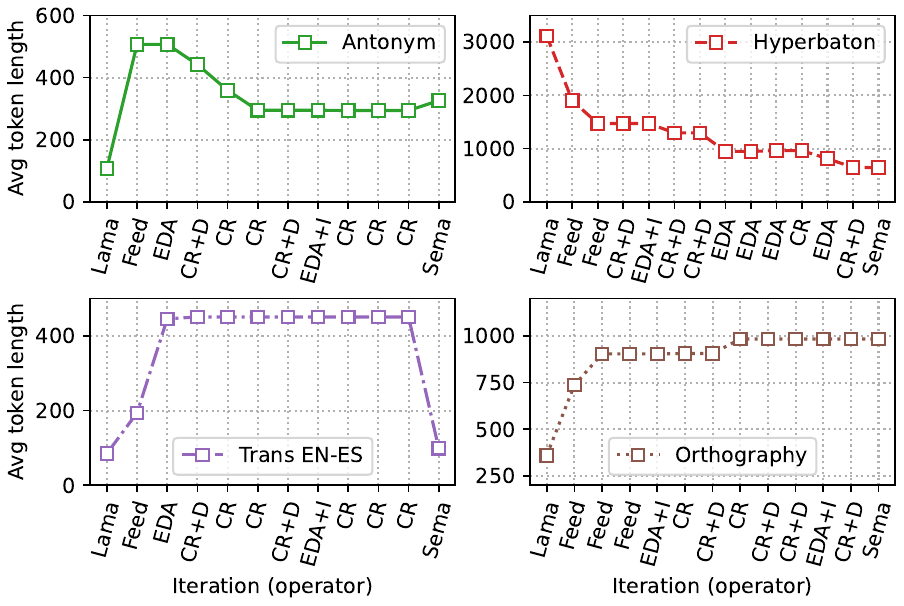}
    \caption{Average prompt length through optimization.}
    \label{fig:token}

\end{figure}

\paragraph{Effect of Operators on Prompt Length} 
Our method is designed to explore the full prompt space, encompassing both zero-shot and few-shot configurations. Understanding how prompt length varies—and how different operators influence this variation—is critical for interpreting the optimization dynamics. Fig. \ref{fig:token} illustrates the average prompt token length over the course of iterations. Interestingly, the length may increase, decrease, or oscillate, which is consistent with the inherently flexible nature of the optimization process. This behavior supports our design rationale, demonstrating the operators' capacity to both add and remove examples as needed. Such variability is not only expected but also essential for navigating the diverse and unconstrained structure of the prompt space effectively.

\paragraph{Effect of Initialization Strategy} 

The \frameName can accommodate two types of inputs: \inputOutputPair and \exampleInstructionnospace, each bringing its own benefits. When using the \inputOutputPair approach, the initialization is solely based on LLM's generations, resulting in greater diversity. On the other hand, initialization in \exampleInstruction draws upon human-provided example prompts, consequently lacking the diversity that \inputOutputPair offers. However, \exampleInstruction empowers users to introduce prior knowledge without relying on LLM's interpretation, which leads to better performance in more complex tasks such as Dyck Languages, and Logical Five as shown in Table \ref{tab:aelp_datasets}.

\paragraph{Prompt Quality} \frameName generates few-shot prompts for 20 / 24 Instruction Induction tasks and 4 / 8 BBH tasks. For hard tasks, \frameName even integrates with different techniques, such as \textit{COT} for task Logical Deduction Five, or adding \textit{“Let’s think step by step”} for the task Reasoning Colored Objects. Beyond prompt techniques, \frameName also generates prompts that are easier for human understanding and more relevant to the tasks. These validates \frameNamenospace's applicability in diverse cases and interpretability for human verifications. More details on prompt quality can be found in Appendix \ref{sec:prompt-quality} where we compare prompts generated by different baselines. All generated prompts are in Section \ref{sec:generated_prompt}.

\paragraph{Hyperparameters}
 \frameName has some hyperparameters such as the threshold for phase transition, and pool size. To test the universal applicability of these settings, we have utilized a threshold of 1\% and a pool size of 15 for initialization, and 5 for the rest of the phases in all 35 tasks. \frameName achieves superior results without specific parameter calibration. The experiments conducted on 7 other models shown in Table \ref{tab:bbh-models} with the same configuration also provide competitive results. Given the superior results in the universal setting, we believe \frameNamenospace\ requires little to no parameter tuning for practical application.

\paragraph{Computational Cost} We evaluate computational cost using two metrics: (1) the total number of API calls to the LLM, and (2) the total token consumption during the end-to-end optimization process. These metrics include both operator application and candidate evaluation steps. We intentionally select these metrics because they directly correlate with the overall runtime and computational overhead of the optimization process.

\begin{figure}[h!]
\centering
    \includegraphics[width=.95\linewidth]{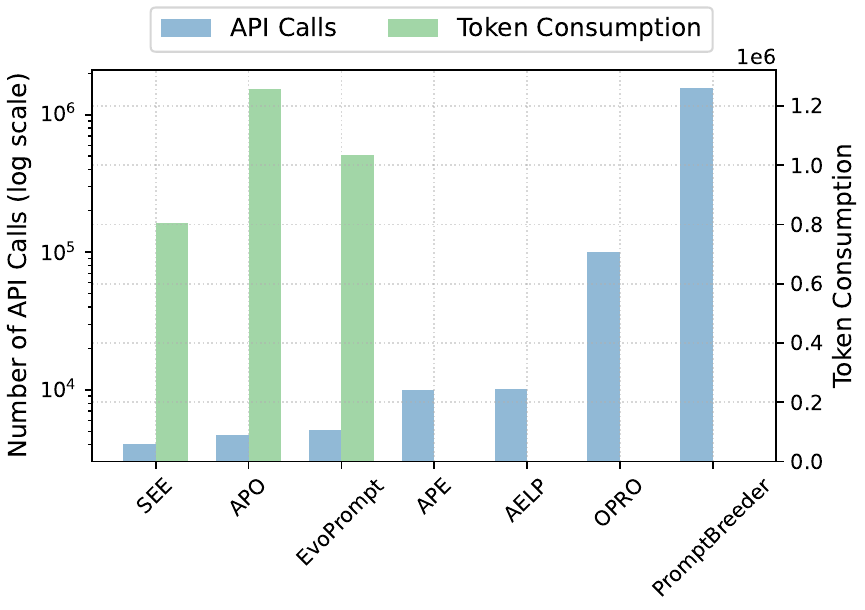}
    \caption{Comparison of computational cost measured by both total API calls and overall token consumption.}
    \label{fig:computational_efficiency}
\end{figure}

As illustrated in Fig. \ref{fig:computational_efficiency}, \frameName demonstrates the highest cost-efficiency, achieving reductions in computational cost by several orders of magnitude compared to other optimization strategies, including those based on metaheuristic approaches. For instance, PromptBreeder—an evolutionary algorithm representing a traditional metaheuristic method—requires approximately 2.5 orders of magnitude more API calls than \frameName.

Given that \frameName, APO, and EvoPrompt exhibit the lowest number of API calls, we further compare these methods based on token consumption on the BBH task formal fallacies. Even under this stricter metric, \frameName remains the most efficient, reinforcing the advantage of its quad-phased design and adaptive operator selection. This innovation significantly enhances the computational efficiency of metaheuristic-inspired optimization frameworks.

Methods such as ZOPO, MoP, and EASE involve additional computational components (e.g., model training or clustering), and are therefore excluded from this analysis to maintain a fair comparison.

\section{Conclusion}
We introduce a cohesive in-context prompt optimization framework that leverages metaheuristic optimization principles to optimize both prompt instructions and examples. Through a strategically designed exploration–exploitation schedule and adaptive operator application, \frameName achieves SOTA performance across a diverse set of benchmark tasks, while significantly reducing computational cost. Notably, \frameName is the first framework to dynamically generate both few-shot and zero-shot prompts, adapting flexibly to the specific requirements of each task.

\section{Limitations}
Although \frameName is the most cost-effective method among baselines, it still needs around $12$ iterations and $4,000$ API calls, which might be insufficient for supporting large-scale online applications. Future work could explore better initialization strategies or data compression techniques to further improve efficiency and reduce latency. Another important opportunity lies in extending the framework beyond single-objective optimization. Developing multi-objective optimization capabilities—such as simultaneously improving accuracy, interpretability, efficiency, or safety—could significantly enhance the framework's flexibility and real-world utility, and is not what \frameName currently focuses on.

\section*{Acknowledgments}
This work includes contributions from Vanderbilt University researchers, supported by funding from Intuit.

\bibliography{custom}

\appendix
\onecolumn

\textbf{Organization}
The Appendix is organized as follows:
\begin{itemize}
    \item \textbf{Appendix~\ref{sec:related_work}. Related Work} \\  
    Related work in automatic prompt optimization.
    
    \item \textbf{Appendix~\ref{sec:operator-def}. Operator Definition} \\  
    We formally define the operators used in our framework.
    
    \item \textbf{Appendix~\ref{sec:operator}. Operator Analysis} \\
    We analyze various properties and behaviors of the operators.
    
    \item \textbf{Appendix~\ref{sec:operator-prompts}. Operator Prompts} \\
    We discuss the prompt templates used for these operators.
    
    \item \textbf{Appendix~\ref{sec:details_exp}. Details of Experiments} \\
    We provide comprehensive details of our experimental setups and protocols.
    
    \item \textbf{Appendix~\ref{sec:add_exp}. Additional Experiment Results} \\
    We present extended results and further analysis beyond the main text.
    
    \item \textbf{Appendix~\ref{fewshot-examples}. Few-shot Add/ Removal Examples} \\
    We illustrate representative showcasing operators' ability to add/ removal examples.
    
    \item \textbf{Appendix~\ref{sec:generated_prompt}. Generated Prompts} \\
    We show all the final prompts automatically generated by our system.
\end{itemize}

\section{Related Work}
\label{sec:related_work}
 
In-context prompting is an efficient approach for communicating LLMs but the performance is strongly affected by the design of the prompt in specifized tasks. Prompt optimization to find the optimal prompt has thus obtained broader attention.
One research direction is the continuous prompt approaches that tune embeddings of input tokens to generate better prompts \cite{li2021prefix, zhang2021differentiable, sun2022black, sun2022bbtv2, Chen2023InstuructZero}. However, the optimized ``soft'' prompts from this paradigm often fall short of interpretability and are inaccessible for blackbox APIs. Discrete prompt approaches \cite{diao2022black, prasad2022grips}, operating discrete tokens directly, offer an interactive interface to humans with better interpretability and show promising performance. Various methods have been proposed via gradient-based search \cite{shin2020autoprompt}, reinforcement learning \cite{zhang2022tempera,deng2022rlprompt, sun2023query} and ensemble methods \cite{hou2023promptboosting,pitis2023boosted}  while these methods encounter concerns in terms of scalability, reliability and efficiency \cite{wang2023promptagent}.  

More recent advancements rely on iterative sampling, scoring, and selection of exceptionally promising prompts, generating diverse possibilities for prompt optimization. \citet{Fernando2023PromptBreeder, Guo2023EVOPrompt, Hsieh2023AELP} proposed leveraging LLMs to implement optimization strategies in prompt searches. \citet{Yang2023OPRO} demonstrates the capability of LLM as optimizers in prompt design. \citet{pryzant2023APO, Zhou2023APE} utilizes natural language feedback to refine prompt instructions. However, these prompt optimization/refinement strategies largely focus on prompt instructions, typically short sentences or paragraphs. While previous search and sampling algorithms have been investigated, such as Monte Carlo search \cite{Zhou2023APE}, Gibbs sampling \cite{xu2023reprompting}, or Beam search \cite{pryzant2023APO}, we introduce a novel dual exploration-exploitation strategy that leverages the in-depth traits of each operator, implementing the metaheuristic optimization framework with adaptive operator selection to enhance the interactive dynamics during optimization. 

\clearpage

\section{Operator Definition}  \label{sec:operator-def}
Operators are used to generate new candidates. Seven types of operators, broadly categorized into five classes are used by \frameNamenospace. The idea is to provide a diverse set of operators so that a broad cognitive space of linguistics is covered. Table \ref{tab:operator-comparison} lists the operators that different kinds of methods use.

\subsection{Lamarckian Operator} \label{subsec:lam}
The Lamarckian operator follows the principles proposed in APE and Prompt Breeder \cite{Zhou2023APE, Fernando2023PromptBreeder}. Given a set of input-output pairs for the task, an LLM agent is used to reverse-engineer the prompt from the provided demonstrating pairs. This type of operator allows a diverse set of prompt candidates to be generated with no prior knowledge of the task. Any prompt candidate will have to be induced from the demonstrating pairs. The prompt used by the LLM agent is in Table \ref{tab:initialPrompt}.

\begin{definition}
\label{def:lam}
(Lamarckian Operator) Given a set of input/output pairs $\mathcal{(Q,A)} = [(Q_1, A_1), ...,(Q_m, A_m)]$ and a base LLM $\mathcal{L}$, the Lamarckian operator is to reverse engineer the instruction $\mathcal{O}_{L}$ so that $\mathcal{O}_{L}(Q_i) = A_i, i=1,...,m$.
\end{definition}

\subsection{Feedback Operator}
Inspired by the concept of \textit{Gradient Descent} in machine learning model training, we introduce an LLM agent that works as an examiner which examines the cases where the current task prompt fails and provides improvement guidance. Such guidance will be treated as \textit{gradient} and be used by another LLM Agent as an improver to generate a new candidate. Though similar to what is proposed in APO \cite{pryzant2023APO}, instead of only using gradient descent repeatedly, which has a higher probability of arriving at a local minimum, we take advantage of its fast converge rate to local minimum and combine it with other operators to target global minimum. When applying the Feedback operator, it will be applied to every candidate in the current pool. The prompt can be found in Table \ref{tab:gradient-generation} - \ref{tab:gradient-application}.

\begin{definition}
\label{def:feedback2}
(Feedback Operator) The Feedback operator generates a new prompt $p^{\prime}$ based on the existing prompt $p \in \mathcal{P}$, and where $p$ made mistakes for a task. The feedback operator $\mathcal{O}_{F}$ first looks at the cases where the current $p$ failed to generate a list of advice $G$, and then asks LLM $\mathcal{L}$ to apply such advice $G$ to existing prompt $p$ for generating the new prompt $p^{\prime}$.
\end{definition}

\subsection{ESTIMATION OF DISTRIBUTION Operator}
The next class of operators takes a set of parents as input to generate a modified candidate. 

\textbf{Estimation of Distribution Operator (EDA)}: Following the principles proposed by \cite{Hauschild2011EDA} and work in \cite{Fernando2023PromptBreeder}, we use a LLM agent that is fed with a subset of the current pool to generate new candidate. To ensure the diversity and quality of the subset, we first rank the candidates in the current pool by their performance in descending order. Then starting from the first item in the ordered candidates, we only add the candidate to the subset if it does not have a similarity score over a threshold with any other candidate that is already in the subset. This way candidates with higher performance are more prone to be added to the subset and the diversity of the subset is achieved. More details on how similarity is calculated can be found in section \ref{sec-sim}. The subset will be randomized before feeding into the LLM agent so the candidate's performance does not dictate its order. The prompt can be found in Table \ref{tab:eda}.

\textbf{EDA and Index Operator}: This is a variant of the EDA operator above. Based on the observations that LLM is more prone to use examples that appear late in the in-context learning \cite{Liu2023Lost, Fernando2023PromptBreeder}, after generating the subset following procedures of EDA, the subset is ordered by their performance in \textit{ascending order}. To further balance exploitation and exploration and avoid being too biased over the candidate with the highest performance \cite{Fernando2023PromptBreeder}, we instructed LLM that the candidates are ranked by their performance in \textit{descending order} so that the low performance candidates are taken into consideration. The prompt can be found in Table \ref{tab:eda+index}.

\begin{definition}
\label{def:eda}
(Estimation of Distribution Operator - EDA) EDA generates a new candidate based on a list of parents. It is a function operator $\mathcal{O}_E$ that performs  $\mathcal{O}_E(\mathcal{P}, $$\mathcal{L}$$)=p^{\prime}$. Given a list of prompts $\mathcal{P} = [p_1,..., p_m]$ and an LLM $\mathcal{L}$, EDA provides a new prompt $p^{\prime}$. Items in $\mathcal P$ satisfy the restriction that $d(p_i, p_j) < t$, where $d$ is a function that calculates similarity, and $t$ is a predefined threshold. If the items in $\mathcal P$ are ordered based on certain criteria, we call it EDA + Index (EDA+I).
\end{definition}

\subsection{Crossover Operator}
This class of operators takes two parents as input to generate a crossover candidate. The prompt can be found in Table \ref{tab:cross over}.

\textbf{Crossover Operator(CR)}: Following the concept of crossover in the optimization algorithm, we introduce an LLM agent to function as a crossover operator that takes two parents and generates a crossover candidate. It takes the best two candidates in the current pool, namely the top two candidates with the highest performance, and performs linguistic crossover.

\textbf{Crossover with Diversity Operator(CR+D)}: This is a variance of the Crossover Operator. To provoke exploration, we follow a similar process in EDA where diversity in parents is considered. Thus it takes the best candidate and the most distinct individual to it as two parents for crossover operation. The distinctness between two candidates is measured by a similarity score. More details on how the similarity score is calculated can be found in section \ref{sec-sim}.

\begin{definition}
\label{def:inj}
(Crossover Operator - CR) Crossover generates a new candidate based on two parents. It is a function operator $\mathcal{O}_{C}$ that performs  $\mathcal{O}_{C}(p_1, p_2, \mathcal{L})=p^{\prime}$ where $p_1, p_2$ are two prompts selected from a prompt pool $\mathcal P$ where $\mathcal{P} = [p_1...,p_m]$, $p^{\prime}$ is the generated prompt that hold features from both $p_1$ and $p_2$.  If $p_2=\argmin_{p \in \mathcal{P}} d(p_1, p_i)$ is applied for choosing $p_2$, we call it Crossover + Distinct (CR + D).
\end{definition}

\subsection{Semantic Operator} \label{subsec:semantic}
This class of operators takes a candidate and uses an LLM agent to compose a new candidate that shares its semantic meaning. When applying the Semantic operator, it will be applied to every candidate in the current pool. The prompt can be found in Table \ref{tab:semantic}.

\begin{definition}
\label{def:seman}
(Semantic Operator) The Semantic operator is a function operator $\mathcal{O}_{S}$ that performs $\mathcal{O}_{S}(p, \mathcal{L})=p^{\prime}$ where $p{\prime}$ is the generated prompt that shares the same semantic meaning as $p$.
\end{definition}

\begin{table}[t]
\centering

\begin{tabular}{@{}c|ccccccc@{}}
\toprule
Operator           & OPRO & EvoPrompt & AELP & PromptBreeder & APE & APO & \frameName \\ 
\midrule
Lamarckian Operator & \cmark & \cmark & - & \cmark & \cmark & - & \cmark \\
Feedback Operator   & - & - & - & - & - & \cmark & \cmark \\
Crossover Operator  & - & \cmark & - & \cmark & - & - & \cmark \\
EDA Operator        & \cmark & - & - & \cmark & - & - & \cmark \\
Semantic Operator   & - & \cmark & \cmark & \cmark & \cmark & \cmark & \cmark \\ 
\bottomrule
\end{tabular}
\caption{Comparison of operators across methods.}
\label{tab:operator-comparison}
\end{table}

\section{Operator Analysis} \label{sec:operator}
\subsection{Few-Shot Additional / Removal Analysis}
\label{sec:op_fewshow}
The ability to add and remove few-shot examples is crucial in dictating whether it is possible to traverse the full in-context prompt space. Thus we conduct the few-shot addition/ removal analysis to gauge each operator's capabilities. Below are examples of how operators add or remove few-shot examples.

\begin{table}[h]
\centering
\begin{center}
\begin{tabular}{ m{2cm} | m{13cm}}
\toprule
 \textbf{Operator Input} & I gave a friend an instruction and some examples The friend read the instruction and wrote an output for every one of the inputs. \newline
Here are the input-output pairs:\newline
\newline
\#\# Example \#\# \newline
\#\# Input \#\# : 92 24  \newline
\#\# Output \#\#: ['68']  \newline

 The instruction was:
 \\ 
 \midrule
 \textbf{Operator Output} & Subtract the second number from the first number and write the result. If the result is negative, write the absolute value of the result. \newline
\newline
Here are the input-output pairs:\newline
\newline
Input: 10 5\newline
Output: ['5']\newline
\newline
Input: 50 75\newline
Output: ['25']\newline
\newline
Input: 100 100\newline
Output: ['0']\newline
\newline
Input: 7 20\newline
Output: ['13']\newline
\newline
Input: 0 5\newline
Output: ['5']\newline
\newline
Input: -10 5\newline
Output: ['15']\newline
\newline
Input: 92 24\newline
Output: ['68']\newline
\newline
Input: 50 -75\newline
Output: ['125']
\\
\bottomrule
\end{tabular}
\end{center}
\caption{Lamarckian Operator Add Few-shot Example}
\label{tab:lam-fewshot}
\end{table}

\begin{table}[h]
\centering
\begin{center}
\begin{tabular}{ m{2cm} | m{13cm}}
\toprule
 \textbf{Original Prompt} & For each word, find its antonym and write it as the output.

 \\ 
 \midrule
 \textbf{Generated Feedback} & \hlmeg{1. Clarify the definition of "antonym" to ensure that the prompt is asking for the opposite meaning of the word, rather than just a different word that is related in some way.} \newline
 \newline
\hlgreen{2. Provide more specific instructions for determining the antonym, such as indicating whether the opposite meaning should be an adjective or a noun.}\newline
\newline
\hlblue{3. Add a list of common antonyms for each word to help users who may not be familiar with all possible opposites.}\newline
\newline
\hllime{4. Provide a context or sentence for each word to help users determine the appropriate antonym.}
 \\ 
 \midrule
  \textbf{Generated Prompt} & For each word, find its antonym, \hlmeg{which is a word that has the opposite meaning, and write it as the output.} \hlgreen{The antonym should be a word that is commonly used to express the opposite meaning of the given word, and it should be of the same part of speech (e.g. noun, verb, adjective, adverb).} \newline
\newline
\hlblue{To help you identify the antonym, we have provided a list of common antonyms for each word. Additionally, we have included a context or sentence for each word to help you determine the appropriate antonym. \\
\\
Example: \\
Word: Hot \\
Antonym: Cold \\}
\hllime{Context: The soup is too hot to eat. \\
\\
Please note that some words may have multiple antonyms depending on the context, so use your best judgment to determine the most appropriate one.}
\\
\bottomrule
\end{tabular}
\end{center}
\caption{Feedback Operator Add Few-shot Example: In this operation, few-shot examples are added based on the feedback. Individual feedback and their corresponding changes are color-coded.}
\label{tab:feedback-fewshot-add}
\end{table}

\begin{table}[h]
\centering
\begin{center}
\begin{tabular}{ m{2cm} | m{13cm}}
\toprule
 \textbf{Original Prompt} & For each word, remove the prefix or suffix that indicates a negative or opposite meaning, and write the resulting word as the output. \newline
 \newline
For example, for the input "destabilize," remove the prefix "de-" which indicates a negative meaning, and the resulting word is "stabilize

 \\ 
 \midrule
 \textbf{Generated Feedback} & \hlmeg{1.Clarify the prompt to specify that it only applies to words with a negative or opposite prefix or suffix. This will prevent confusion and mistakes when dealing with words that do not have such affixes.} \newline
 \newline
\hlgreen{2. Provide examples of common negative or opposite prefixes and suffixes, such as "un-" or "-less," to help users identify them more easily.}\newline
\newline
\hlblue{3. Include a check to ensure that the resulting word is a valid English word, to avoid outputs like "ane" or "direct" that are not actual words. This could be done by cross-referencing the output with a dictionary or word list.}\newline
\newline
\hllime{4. Consider adding an option to retain the original word if no negative or opposite prefix or suffix is present. This would eliminate confusion when dealing with words like "linear" that do not have such affixes.}
 \\ 
 \midrule
  \textbf{Generated Prompt} & For each word that contains a negative or opposite prefix or suffix, remove the affix and write the resulting word as the output. \hlgreen{Common negative or opposite prefixes and suffixes include "un-," "dis-," "in-," "-less," and "-ful."} \hlmeg{Please note that this prompt only applies to words with these specific affixes. } \newline
\newline
\hlblue{To ensure that the resulting word is a valid English word, we will cross-reference the output with a dictionary or word list. If the resulting word is not a valid English word, it will be excluded from the output.} \newline
\newline
\hllime{If the original word does not contain a negative or opposite prefix or suffix, you may choose to retain the original word or exclude it from the output. Please specify your preference in the input.}

\\
\bottomrule
\end{tabular}
\end{center}
\caption{Feedback Operator Remove Few-shot Example: In this operation, few-shot examples are removed based on the feedback. Individual feedback and their corresponding changes are color-coded.}
\label{tab:feedback-fewshot-remove}
\end{table}

\begin{table}[h]
\centering
\begin{center}
\begin{tabular}{ m{1.5cm} | m{13.5cm}}
\toprule
 \textbf{Operator Input} & Order adjectives correctly in English sentences. \newline
 \newline
Q: Which sentence has the correct adjective order: \newline
Options: \newline
(A) rubber terrible ship 
(B) terrible rubber ship \newline
A: Let's think step by step. 
When there is more than one adjective before a noun, the adjectives need to respect the following order before a noun: "[1. opinion] [2. size] [3. age] [4. shape] [5. color] [6. origin] [7. material] [8. purpose] noun".
Option (A): "rubber terrible ship". (1) rubber" falls into the material category. (2) "terrible" falls into the opinion category. Option (A) has the following adjective order: [7. material] [1. opinion] (or, in numeric terms, 7 1). Because 7 < 1 is not correct, (A) does not have the correct ordering.
Option (B): "terrible rubber ship". Option (B) has the following adjective order: [1. opinion] [7. material] (or, in numeric terms, 1 7). Because 1 < 7 is correct, (B) has the correct ordering. So the answer is (B). \newline
 \newline
Q: Which sentence has the correct adjective order: \newline
Options: \newline
(A) repulsive small Brazilian exercise ship 
(B) Brazilian repulsive exercise small ship \newline
A: Let's think step by step. 
When there is more than one adjective before a noun, the adjectives need to respect the following order before a noun: "[1. opinion] [2. size] [3. age] [4. shape] [5. color] [6. origin] [7. material] [8. purpose] noun".
Option (A): "repulsive small Brazilian exercise ship". (1) "repulsive" falls into the opinion category. (2) "small" falls into the size category. (3) "Brazilian" falls into the origin category. (4) "exercise" falls into the purpose category. Option (A) has the following adjective order: [1. opinion] [2. size] [6. origin] [8. purpose] (or, in numeric terms, 1 2 6 8). Because 1 < 2 < 6 < 8 is correct, (A) has the correct ordering.
Option (B): "Brazilian repulsive exercise small ship". Option (B) has the following adjective order: [6. origin] [1. opinion] [8. purpose] [2. size] (or, in numeric terms, 6 1 8 2). Because 6 < 1 < 8 < 2 is not correct, (B) does not have the correct ordering. So the answer is (A). \newline
 \newline
Q: Which sentence has the correct adjective order: \newline
Options: \newline
(A) blue gold wonderful square shoe 
(B) wonderful square blue gold shoe \newline
A: Let's think step by step. 
When there is more than one adjective before a noun, the adjectives need to respect the following order before a noun: "[1. opinion] [2. size] [3. age] [4. shape] [5. color] [6. origin] [7. material] [8. purpose] noun".
Option (A): "blue gold wonderful square shoe". (1) "blue" falls into the color category. (2) "gold" falls into the material category. (3) "wonderful" falls into the opinion category. (4) "square" falls into the shape category. The adjective order that Option (A) has is [5. color] [7. material] [1. opinion] [4. shape] (or, in numeric terms, 5 7 1 4). Because 5 < 7 < 1 < 4 is not correct, (A) does not have the correct ordering.
Option (B): "wonderful square blue gold shoe". Option (B) has the following adjective order: [1. opinion] [4. shape] [5. color] [7. material] (or, in numeric terms, 1 4 5 7 ). Because 1 < 4 < 5 < 7 is correct, (B) has the correct ordering. So the answer is (B).

 \\ 
 \midrule
 \textbf{Operator Output} & Rearrange the adjectives in the given sentence in the correct order.

\\
\bottomrule
\end{tabular}
\end{center}
\caption{Semantic Operator Remove Few-shot Example}
\label{tab:seman-fewshot}
\end{table}

\clearpage

\subsection{Operator Feature Analysis}
\label{sec:op-feature}
To study the features of each operator we conduct a preliminary experiment where we study four operators: EDA Operator, Crossover, Feedback Operator, and Semantic Operator. 

\paragraph{Initialization:} As the initialized points have a tremendous impact on optimization problems. We randomly use four different seeds to create four initial pools for four different tasks: Causal Judgement,  Salient Translation Error Detection, Disambiguation QA, and Hyperbaton. The idea is to provide various initialization points so that the performance of operators can be averaged to rule out the influence of initialization.

\paragraph{Operator Applications:} For each initialization, we apply the following procedure for all four operators. 
\begin{itemize}
    \item For one round, starting with the initial pool, we consecutively apply the operator 5 times. This is to study the value of applying the operator consecutively.
    \begin{itemize}
        \item For EDA and CrossOver, as they require multiple parents, we keep a pool size of 5 for each iteration after applying the operator. Performance gain is defined as whether the average performance of the pool is improved.
        \item For Feedback Operator and Semantic Operator, as they only need one parent, we apply them to a random candidate from the initial pool and use the new candidate as the base for the next round. Performance gain is defined as whether the new candidate has a higher performance than its parent.
    \end{itemize}
    \item To reduce the impact of randomness, we run this process 5 rounds for each operator.
\end{itemize}
Thus for each operator, it will be run a total of 4 tasks * 5 rounds * 5 application = 100 times.

\begin{figure}[h!]
\centering
    \includegraphics[width=0.4\textwidth]{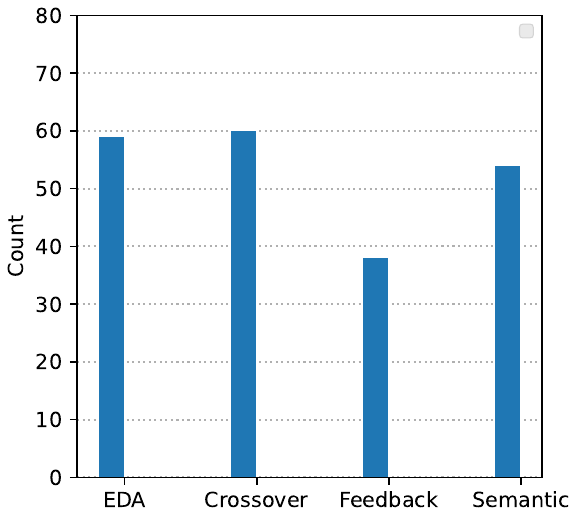}
    \caption{Operator Improvement Count}
    \label{fig:operator_count}
    \vspace{-10pt}

\end{figure}

\paragraph{Analysis:} There are two aspects we are particularly interested in. The first is \textbf{what the likelihood of performance gain when applying an operator is} (Probability of Improvement), and the second is \textbf{how fast each operator can continuously bring improvement} (Convergence Speed).

\begin{figure}[h!]
\centering
    \includegraphics[width=0.9\textwidth]{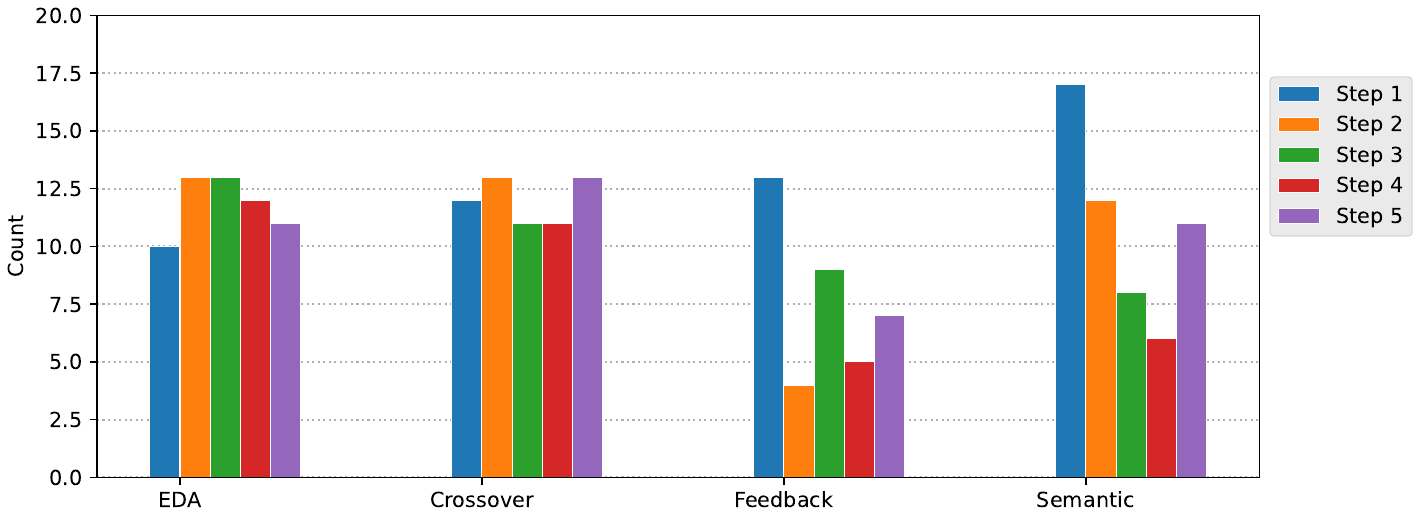}

    \caption{Operator Improvement Pattern: EDA Operator and Crossover have similar improvement counts for each step whereas for Feedback Operator and Semantic Operator, the first step introduced significantly more times of improvement compared to the others.}
    \label{fig:operator_pattern}
    \vspace{-10pt}

\end{figure}

\begin{figure}[h!]
\centering
     \includegraphics[width=0.7\textwidth]{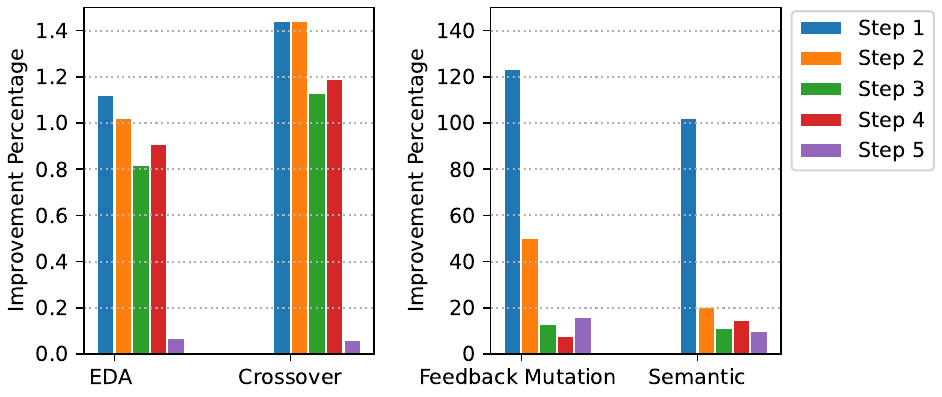}
    \caption{Improvement Ratio: On the left, for EDA and Crossover, we observe an almost equal improvement ratio for the first four steps. Improvement Ratio is defined as the relative percentage of improvement in the average performance for the entire pool. On the right, for Feedback and Semantic Operator, we observe the first round contributes significantly more improvement compared to the others. As Feedback and Semantic Operators take one input candidate, Improvement Ratio is defined as the relative performance improvement percentage for the candidate after applying the operator.}
    \label{fig:operator_ratio}
    \vspace{-10pt}

\end{figure}

\begin{itemize}
    \item \textbf{Probability Of Improvement}: Figure \ref{fig:operator_count} shows the number of times performance is improved by each operator. Crossover and EDA Operator introduces improvements in more steps with Semantic Operator ranking third. Feedback Operator introduces the least number of improvements. This result helps populate the \textit{Prob} column in table \ref{tab:operator}.
    \item \textbf{Convergence Speed}: Figure \ref{fig:operator_pattern} shows that for each operator, as they are applied in 5 consecutive steps, the number of times improvement is introduced for each step. Figure \ref{fig:operator_ratio} shows the average percentage of performance gain operators brought in each step.

    \begin{itemize}
        \item For EDA Operator and Crossover, each 5 step has a similar number of contributions for performance gains as shown in figure \ref{fig:operator_pattern}. From figure \ref{fig:operator_ratio} we can also observe the first step brings the most improvement and the first 4 steps bring a similar improvement ratio.
        \item For Feedback Operator and Semantic Operator, the first step has a significantly higher chance of introducing improvement as shown in figure \ref{fig:operator_pattern}. This is especially true for Feedback Operator where step 1 accounts for over 34\% of the total improvement counts. As for the improvement ratio, the first step for both Feedback Operator and Semantic Operator introduces significantly more improvements than the rest of the steps shown in figure \ref{fig:operator_ratio}.
    \end{itemize}

Based on the tests, we learned that the value gained for applying Feedback Operator and Semantic Operator is significantly reduced after the 1st application. We interpret it as \textbf{Feedback Operator and Semantic Operator can jump to the local minimum pretty fast}, namely in 1 step, thus leading to less possibility of improvement for steps 2 - 5. Whereas for EDA Operator and Crossover, as they are merging genetic information between candidates, the likelihood of improvement is relatively randomized. So even if the first round of applying them renders no improvement, there is still a chance of performance gain in the following run. In other words, \textbf{we should be more patient with EDA Operator and Crossover}. Thus the operator tolerance (described in section \ref{sec:stop-criteria}-design 2) for EDA and Crossover is set to 4 and for Feedback Operator and Semantic Operator is 1. These learnings help populate the \textit{Speed} column in table \ref{tab:operator}. 
\end{itemize}

\clearpage

\section{Operator Prompts}
\label{sec:operator-prompts}

\paragraph{Operator Implementation:} The state-of-art frameworks such as APO, EVOPROMPT, and AELP have already implemented operators such as feedback operator, crossover operator, and semantic operator with LLM. However, these implementations inflict restrictions on LLM with prompts. For example, in APO when implementing the feedback operator, the prompt specifically identified the use case to be zero-shot. \cite{pryzant2023APO} In EVOPROMPT-DE, when applying crossover operators, the focus is to only change the parts that two parents differentiate from each other. \cite{Guo2023EVOPrompt} In AELP, when applying semantic operators, it is restricted to a sentence level, not the whole prompt. \cite{Hsieh2023AELP}. In \frameNamenospace, we pay special attention not to apply any restrictions in our operator prompt, realizing the full potential of LLMs.

\begin{table}[h]
\centering
\begin{tabular}{p{0.8\textwidth}}
\toprule 
I gave a friend an instruction and some input. The friend read the instruction and wrote an output for every one of the inputs.
Here are the input-output pairs:
\\ \\
\#\# Example \#\#  \\
\{\textit{input output pairs}\} 

\\The instruction was:
\\
\bottomrule
\end{tabular}
\caption{Lamarckian Operator Prompt}
\label{tab:initialPrompt}
\end{table}

\begin{table}[h]
\centering
\begin{tabular}{p{0.8\textwidth}}
\toprule
You are a quick improver. Given an existing prompt and a series of cases where it made mistakes. Look through each case carefully and identify what is causing the mistakes. Based on these observations, output ways to improve the prompts based on the mistakes. \\
\\
\#\# \hl{Existing Prompt} \#\#  \\
\{\textit{existing prompt}\} \\
\\
\#\# Cases where it gets wrong:\#\#  \\
\{\textit{wrong cases}\} \\
\\
\hl{ways to improve} the existing prompt based on observations of the mistakes in the cases above are:
\\
\bottomrule
\end{tabular}
\caption{Gradient Descent Generation Prompt: Unlike APO which is also using gradient descent, we are \hl{NOT adding restrictions} such as \textit{"zero-shot classifier prompt."}, \hl{nor providing any differentiation} between \textit{instructions} and \textit{examples}. Instead, we specifically ask LLM to output multiple feedback in one go. Also as are \hl{passing in the existing prompt as a whole}, thus feedback should be on the paragraph/prompt level instead of the sentence/instruction level. We highlight the design that helps us achieve this below.}
\label{tab:gradient-generation}
\end{table}

\begin{table}[h]
\centering
\begin{tabular}{p{0.8\textwidth}}
\toprule
You are a quick improver. Given an existing prompt and feedback on how it should improve. Create an improved version based on the feedback.
\\
\\
\#\# \hl{Existing Prompt} \#\#  \\
\{\textit{existing prompt}\} \\
\\
\#\# Feedback\#\#  \\
\{\textit{feedback}\} \\
\\
\#\# \hl{Improved Prompt}\#\# \\

\bottomrule
\end{tabular}
\caption{Gradient Descent Application Prompt: Following the principle of optimizing prompt as a whole, our operator prompts take input and output on the entire prompt level}
\label{tab:gradient-application}
\end{table}

\begin{table}[h]
\centering
\begin{tabular}{p{0.8\textwidth}}
\toprule
You are a mutator. Given a series of prompts, your task is to generate another prompt with the same semantic meaning and intentions.

\\
\\
\#\# Existing Prompts \#\#  \\
\{\textit{existing prompt}\} \\
\\

The newly mutated prompt is:\\

\bottomrule
\end{tabular}
\caption{EDA Prompt}
\label{tab:eda}
\end{table}

\begin{table}[h]
\centering
\begin{tabular}{p{13cm}}
\toprule
You are a mutator. Given a series of prompts, your task is to generate another prompt with the same semantic meaning and intentions. 
\hl{The series of prompts are ranked by their quality from best to worst.}

\\
\#\# Existing Prompts \#\#  \\
\{\textit{existing prompt}\} \\
\\

The newly mutated prompt is:\\

\bottomrule
\end{tabular}
\caption{EDA+Index Prompt: The difference between EDA + Index and EDA is that EDA + Index takes advantage of the in-context learning technique and informs the order of the passed-in prompts}
\label{tab:eda+index}
\end{table}

\begin{table}[h]
\centering
\begin{tabular}{p{0.8\textwidth}}
\toprule
You are a mutator who is familiar with the concept of cross-over in genetic algorithm, namely combining the genetic information of two parents to generate new offspring. Given two parent prompts, you will perform a cross-over to generate an offspring prompt that covers the same semantic meaning as both parents.\newline
\newline
\# Example \\
Parent prompt 1: Now you are a categorizer, your mission is to ascertain the sentiment of the provided text, either favorable or unfavorable\newline
\newline
Parent  prompt 2: Assign a sentiment label to the given sentence from ['negative', 'positive'] and return only the label without any other text.\newline
\newline
Offspring prompt: Your mission is to ascertain the sentiment of the provided text and assign a sentiment label from ['negative', 'positive’]. \\
\\
\#\# Given \#\#  \\
Parent prompt 1: \{\textit{prompt 1}\} \\
Parent prompt 2: \{\textit{prompt 2}\} \\
Offspring prompt: \\

\bottomrule
\end{tabular}
\caption{Cross Over Prompt}
\label{tab:cross over}
\end{table}

\begin{table}[h]
\centering
\begin{tabular}{p{0.8\textwidth}}
\toprule
You are a mutator. Given a prompt, your task is to generate another prompt with the \hl{same semantic meaning and intentions.} \\
\\
\# Example: \\
current prompt: Your mission is to ascertain the sentiment of the provided text and assign a sentiment label from ['negative', 'positive’]. \\
mutated prompt:  Determine the sentiment of the given sentence and assign a label from ['negative', 'positive']. \\
\\
\\
Given:\\
current prompt: \{\textit{existing prompt}\} \\
mutated prompt::\\

\bottomrule
\end{tabular}
\caption{Semantic Operator Prompt: To provoke LLM's creativity, we do not restrict to the \hl{semantic} level but expand that to \hl{intentions}, allowing LLM to not \hl{stick to a sentence-by-sentence modification.}}
\label{tab:semantic}
\end{table}
\clearpage

\section{Details of Experiments}
\label{sec:details_exp}

\paragraph{Implementation Details.} We utilized GPT-3.5-turbo to develop LLM agents capable of performing various operators in all tasks. The GPT-3.5-turbo access was through internal hosting of Azure, where additional security guardrails prevented some API calls from completing for security reasons. As such, accuracy was computed only on successful responses. We conduct comparisons between GPT 3.5 and GPT. 4 in 4 BBh tasks and all the other tasks. We set up training, development, and testing datasets, select the prompt with the highest score on the dev set, and report its score on the testing set. For BBH and APO tasks, we conducted three end-to-end runs, with the average performance and standard deviation reported in Table \ref{tab:aelp_avg} and Table \ref{tab:apo_datasets}. For additional parameter settings please refer to Section \ref{sec:setting}.

\subsection{Benchmark tasks}
\label{sec:tasks}
\begin{itemize}[leftmargin=10pt]
    \item {\bf 24 Instruction Induction Tasks}: These 24 instruction tasks \cite{Honovich2022Instuct} span many facets of language understanding, from simple phrase structure to similarity and causality identification. Both training and testing data are provided for these tasks and we create our training and development data set from the available training data and use the provided testing data set as is. Depending on the task, we use up to 50 training data and up to 50 development data. We use \inputOutputPair format for these tasks.
    \item {\bf Ethos}: Ethos \cite{Mollas2021Ethos} is an online English hate speech detection data set with 997 online comments and hate speech labels. We select 50 for training, 50 for development, and 150 for testing. We use \exampleInstruction format for this data set following the practice of APO \cite{pryzant2023APO}.
    \item {\bf Liar}: Liar \cite{Wang2017Liar} is an English fake news detection data set with 4000 statements, context, and lie labels. We select 50 for training, 50 for development, and 150 for testing. We use \exampleInstruction format for this data set following the practice of APO \cite{pryzant2023APO}.
    \item {\bf Sarcasm}: Sarcasm \cite{Farha2020Sar} is an Arabic sarcasm detection data set with 10,000 online comments and sarcasm labels. We select 50 for training, 50 for development, and 150 for testing. We use \exampleInstruction format for this data set following the practice of APO \cite{pryzant2023APO}.
    \item {\bf BBH}: BBH \cite{srivastava2023beyond} is a collaborative benchmark that aims to quantitatively measure the capabilities and limitations of language models. We followed the same practice in the AELP paper with the same tasks and randomly selected 125 for training/ development, and up to 125 for testing. \cite{Hsieh2023AELP}
\end{itemize}

\subsection{Baselines}
\label{sec:baselines}
\paragraph{9 Baselines.} We evaluate \frameName against a variety of LLM-based approaches that have achieved state-of-the-art performance in prompt optimization:   

\begin{itemize}[leftmargin=10pt, itemsep=0.5pt, topsep=1.5pt, partopsep=1.5pt]
    \item {\bf APE} \cite{Zhou2023APE}, {\bf ZOPO} \cite{huw2024zopo} and {\bf APO} \cite{pryzant2023APO}: APE utilizes a Monte Carlo Search strategy that emphasizes {\em exploration}, while APO emphasizes {\em exploitation}, which harnesses incorrect instances as feedback gradient. ZOPO utilizes zeroth-order optimization methods to find local optimal.
    \item {\bf OPRO} \cite{Yang2023OPRO}: OPRO leverages LLM as optimizers to generate better instruction via meta-prompt, solution-score pairs, and task descriptions. 
    \item {\bf PromptBreeder} \cite{Fernando2023PromptBreeder}, {\bf EvoPrompt} \cite{Guo2023EVOPrompt} and {\bf AELP} \cite{Hsieh2023AELP}: these methods connect LLMs with evolution algorithms for prompt optimization. 
    \item {\bf MoP} \cite{wang2024mop}, {\bf EASE} \cite{wu2024ease}: these methods can optimize instructions and examples simultaneously.

\end{itemize}

\subsection{\frameName Setting}
\label{sec:setting}
\begin{itemize} [leftmargin=10pt]
\item {\bf Pool Size}: In the experiments, for \textit{phase 0: Global initialization} we 
set the pool size to be 15. For the rest phases, we set the pool to be 5.

\item {\bf Operator Tolerance}: Based on operator analysis in section \ref{sec:op-feature}, the tolerance for Feedback Operator and Semantic Operator is set to 1. The tolerance for EDA Operator and Crossover is set to 4. Thus the minimum number of times operators will be applied in \textit{phase 2: global optimization operation} is 8.

\item {\bf Model Configuration}: For operators, we set the temperature to 0.5 to tap into LLM's creativity. For performance evaluations, we set the temperature to 0.

\item {\bf Performance Gain in Stop Criteria}: To improve efficiency, when evaluating performance gain to decide whether we should move to the next phase, we are only looking at the best candidate in the current pool.

\item {\bf Candidate Selection}: To improve efficiency, after getting new candidates, we combine them with the current pool and use a greedy algorithm to select the top performer to be the new pool.
\end{itemize}

\clearpage

\section{Additional Experiment Results} \label{sec:add_exp}

\subsection{BBH Task Average \& Standard Deviation}
We run each method three times and report and average and standard deviation in Table \ref{tab:aelp_avg}.

\begin{table*}[h!]
\resizebox{\linewidth}{!}{
\small
\begin{tabular}{@{}l|llllllll@{}}
\toprule
Method       & \begin{tabular}[l]{@{}l@{}}Causal\\ Judgement\end{tabular} &  \begin{tabular}[l]{@{}l@{}}Dis\\-ambiguation\end{tabular}  & \begin{tabular}[l]{@{}l@{}}Dyck \\ Languages\end{tabular} & \begin{tabular}[l]{@{}l@{}}Formal \\Fallacies\end{tabular} & Hyperbaton & \begin{tabular}[l]{@{}l@{}}Logical\\ Five\end{tabular} & \begin{tabular}[l]{@{}l@{}}Color\\ Reasoning\end{tabular} & \begin{tabular}[l]{@{}l@{}}Salient \\      Translation\end{tabular} \\ \midrule
\\
\frameNamenospace-pair     
& $69.97_{(2.45)}$	 &${\bf 69.90}_{(3.53)}$	&$7.06_{(1.23)}$	&${\bf 58.49}_{(0.41)}$	&${\bf 84.36}_{(2.24)}$	&$45.49_{(2.73)}$	&$58.13
_{(2.36)}$ &${\bf 48.38}_{(0.81)}$
\\
\frameNamenospace-example  
&${\bf 84.85}_{(5.45)}$	&$68.01_{(0.4)}$	&${\bf 35.48}_{(12.18)}$	&$53.06_{(4.95)}$	&$81.58_{(9.89)}$	&${\bf 73.56}_{(8.99)}$	&${\bf 77.15}_{(4.13)}$	&$47.01_{(0.88)}$
\\
\bottomrule
\end{tabular}
}
\caption{BBH Tasks Average and Standard Deviation}
\label{tab:aelp_avg}
\end{table*}

\subsection{3 Detect Task for APO}
Below are the results of \frameName on 3 detection task compared with APO. 
\begin{table}[h!]
\centering
\begin{tabular}{@{}l|lll@{}}
\toprule
Method  & Ethos & Liar & Sarcasm \\ \midrule
APO \cite{pryzant2023APO}     & 0.95  & 0.51 & 0.85    \\
\frameName (GPT-3.5) & $0.96_{(0.96)}$  & $0.61_{(3.85)}$ & $0.87_{(1.25)}$    \\
\frameName (GPT-4)   & 0.96  & 0.69 & 0.89    \\ \bottomrule
\end{tabular}
\caption{Testing performance on 3 detect tasks from APO.}
\label{tab:apo_datasets}
\end{table}

\subsection{24 Instruction Induction Tasks}
\label{sub:apetask}
Table \ref{tab:apetask} shows the comparison between APE, PromptBreeder, MoP, EvoPrompt, OPRO, EASE, ZOPO and \frameName evaluated by the best prompt on 24 instruction induction tasks. For EASE we use the results with instruction for a fair comparison. For ZOPO, we use the better performance between the two versions. 

\frameName outperforms 23 / 24 tasks over APE zero shot, 21 / 24 tasks over APE few shot, 22 / 24 tasks over Prompt Breeder, 21 / 24 tasks over MoP, 14 / 14 tasks over EvoPrompt, 14 / 14 tasks over OPRO, 10 / 15 tasks over EASE and 14 / 14 tasks over ZOPO.

\frameName generated few-shot prompts for 20 / 24 tasks and zero-shot examples for 4 / 24 tasks. For the full set of generated prompts please refer to Table \ref{tab:ape-prompt}.

\begin{longtable}[c]{ @{}m{3cm} | m{0.8cm} |m{0.8cm} |m{0.8cm} |m{0.8cm} |m{0.8cm} |m{0.8cm} | m{0.8cm} | m{0.8cm} | m{1cm} | m{1cm}@{}}

 \endfirsthead

 \hline
 \multicolumn{2}{l}{Continuation of Table \ref{tab:apetask}}\\
 \hline

 \endhead

\hline
 \multicolumn{2}{l}{Continuation of Table \ref{tab:apetask}}\\
 \hline

 \endfoot

 \endlastfoot

 \toprule
\textbf{Task} &\textbf{APE} (zero-shot) & \textbf{APE} (few-shot) & \textbf{PB} (few-shot)  & \textbf{MoP} & \textbf{Evo Pro-mpt} & \textbf{OP-RO} & \textbf{EA-SE} (w/in) & \textbf{ZO-PO} (best)& \textbf{\frameNamenospace}-3.5 &\textbf{\frameNamenospace}-4\\
 \midrule
Antonyms & 0.83 & 0.86 & 0.87 &0.88  & 0.84 & 0.79 & 0.85 & 0.85 &   \textbf{0.89} & \textbf{0.91}
\\ \midrule
Cause Effect & 0.84 & 1 & 1  & 0.93 & 0.84 & 0.83 & \_ & 0.95 &   0.96 & \textbf{1}
\\ \midrule
Common Concept  & 0.27 & 0.32 & 0 & \textbf{0.38} &  0.11 & 0.09 & \_ & 0.24 &  0.23 & 0.28
\\ \midrule
Diff & 1& 1 & 1 & 1 & 0.27 & 1 & 1 & 1 & \textbf{1 }  & \textbf{1 }
\\ \midrule
First Word Letter & 1 & 1 & 1& 1 & \_ & \_ & \_ & \_ & \textbf{1 } & \textbf{1 } 
\\ \midrule
Informal Formal  & 0.65 & \textbf{0.70} & 0.07 & 0.63 & 0.52 & 0.48 & \_ & 0.62 &   0.6 & 0.67
\\ \midrule
Large Animal & 0.97 & 0.97 & 0.97 & 0.96 & \_ & \_ & \textbf{1} & \_ &   0.96 & 0.94
\\ \midrule
Letters List  & 0.99 & 1 & 0.99 & 0.99  & 1 & 0.99 & \_ & 1  &  \textbf{1} &  \textbf{1}
\\ \midrule
Taxonomy Animal & 0.66 & 0.79 & 1 &0.72 & 0.83 & 0.30 & 1 & 0.90 &   0.96& \textbf{1 }
\\ \midrule
Negation & 0.83 & 0.9 & 0.9 &  0.87 & 0.86 & 0.73 & \textbf{1} & 0.86  & 0.94 & 0.88
\\ \midrule
Num Verb & 1 & 1 & 1 & 1 & \_ & \_ & \_ & \_ & \textbf{1  }  & \textbf{1  } 
\\ \midrule
Active Passive & 1 & 1 & 1 & 1 & \_ & \_ & \_ & \_ & \textbf{1  }  & \textbf{1  }
\\ \midrule
Singular Plural & 1 & 1 & 1 & 1 & \_ & \_ & \_ & \_ & \textbf{1  }  & \textbf{1  }
\\ \midrule
Rhymes & 1 & 0.61 & 1 & 0.94 & 0.60 & 0.23 & 1 & 1 &  \textbf{1}  & \textbf{1  }
\\ \midrule
Second Word Letter  & 0.87 & 0.69 & 0.95 &0.75 & 0.25 & 0.87 & 1 & 0.97 &  \textbf{1}  & \textbf{1  }
\\ \midrule
Sentence Similarity  & 0.36 & 0.43 & 0.56 &\textbf{0.68} & 0.02 & 0.03 &  0.58 & 0.37 &   0.38 & 0.55
\\ \midrule
Sentiment & 0.94 & 0.93 & 0.93 & 0.97 & \_ & \_ & \textbf{1} &\_ & 0.94 & 0.94
\\ \midrule
Orthography Starts  & 0.68 & 0.69 & 0.71  &0.72 & 0.15 & 0.34 & 0.82 & 0.71 & 0.72 &\textbf{0.94}
\\ \midrule
Sum & 1 & 1 & 1& 1 & 1  & 1  & 1  & 1  & \textbf{1}  & \textbf{1} 
\\ \midrule
Synonym  & 0.22 & 0.14 & 0.43 &0.26 & 0.40 & 0.40 & 0.32 & 0.45 & \textbf{0.46} & 0.38
\\ \midrule
Trans En De  & 0.72 & 0.86 & 0.87 &0.72  & \_  & \_  & 0.90 & \_ &   0.83 &  \textbf{0.96}
\\ \midrule
Trans En Es & 0.86 & 0.91 & 0.91 & 0.86 & \_  & \_  & \textbf{1} & \_  &  0.92 &  0.94
\\ \midrule
Trans En Fr  & 0.78 & 0.9 & 0.91 & 0.79 & \_  & \_  & 0.85 & \_  &   0.88 &  \textbf{0.93}
\\ \midrule
Word in Context  & 0.62 & 0.63 & 0.65 & 0.67  & \_  & \_ & \_  & \_  &  0.66 & \textbf{0.7}
\\ \bottomrule
\caption{24 Instruction Induction Task in APE}
\label{tab:apetask}
 \end{longtable}

 \subsection{Generated Prompt Comparison} \label{sec:prompt-quality}
 We notice that the \textbf{prompts generated by \frameName are easier to understand by humans.} Below is a comparison between prompts generated for task Rhymes. The task description is: \textit{"Write a word that rhymes with the input
word.".,}

 The prompt generated by APE and ZOPO does not fit the task. The prompt generated by Prompt Breeder is not easy to understand how it relates to rhyme. The prompt generated by \frameName is easy to understand with few shot examples added.

  \begin{longtable}[c]{ m{2cm} | m{13cm} }
 \endfirsthead

 \hline
 \multicolumn{2}{l}{Continuation of Table \ref{tab:promptcompare}}\\
 \hline

 \endhead

\hline
 \multicolumn{2}{l}{Continuation of Table \ref{tab:promptcompare}}\\
 \hline

 \endfoot

 \endlastfoot

 \toprule
\textbf{Framework} & \textbf{Generated Prompt} \\
 \midrule
 \textbf{APE } & write a function that takes in a string and outputs the string with the first letter capitalized.
\\ \midrule
 \textbf{Prompt Breeder} & Prompt 0: If the last letter of the input is ’e’, remove it. \newline
Prompt 1: remove the last two letters of the input and add the
letters \textbackslash xc2 \textbackslash x93mote \textbackslash xc2 \textbackslash x94. \newline
Contexts \newline
Context 0: \newline
Q. pea \newline
A. If the last letter of the input is ’e’, remove it. \newline
A. If the last letter of the input is ’s’, remove it. \newline
A. If the last letter of the input is ’y’, remove it. \newline
A. If the last letter of the input is \newline
remove the last two letters of the input and add the letters \textbackslash xc2 \textbackslash
x93mote \textbackslash xc2 \textbackslash x94.\newline
Therefore, the correct answer is (a) pea.\newline
Context 1:\newline
Q. night\newline
A. If the last letter of the input is ’e’, remove it.\newline
A. If the last letter of the input is ’t’, remove it.\newline
A. If the last letter of the input is ’h’, remove it.\newline
A. If the last letter of the input is\newline
remove the last two letters of the input and add the letters \textbackslash xc2 \textbackslash
x93mote \textbackslash xc2 \textbackslash x94.\newline
Therefore, the correct answer is (The answer is night.\newline
Context 2:\newline
Q. add\newline
A. If the last letter of the input is ’e’, remove it.\newline
A. If the last letter of the input is ’d’, remove it.\newline
A. If the last letter of the input is ’a’, remove it.\newline
A. If the last letter of the input is\newline
remove the last two letters of the input and add the letters \textbackslash xc2 \textbackslash
x93mote \textbackslash xc2 \textbackslash x94.\newline
Therefore, the correct answer is (The answer is added.
\\ \midrule
\textbf{ZOPO} & The prompts were to create a program that takes in a word as input and outputs a related word based on a specific set of rules. The rules are as follows: If the input word starts with “tri", the output should be “slip".
\\ \midrule
 \textbf{\frameName}  & Create a list of words that rhyme with the given word. To ensure that your rhymes are accurate, make sure that the words have the same vowel sound and ending consonant sound. For example, "cat" rhymes with "bat" and "hat," but not with "dog" or "mat." \newline
\newline
To get started, here are some examples of words that rhyme with the given word: \newline
\newline
- Love: dove, glove, above, shove, of \newline
- Time: rhyme, chime, climb, mime, prime \newline
\newline
To find more rhyming words, you can use a rhyming dictionary, online resources, or brainstorm with friends. Be creative and try to use a variety of different rhyming words instead of repeating the same one multiple times. \newline
\newline
To avoid common pitfalls, make sure to double-check your spelling and pronunciation of the words. Also, avoid using words that only partially rhyme or have a different stress pattern. \newline
\newline
After you've created your list, ask for feedback on the quality of your rhymes. This can help you to improve and refine your skills. \newline
\newline
For an added challenge, consider generating rhyming words that fit a particular theme or context. This can help you to focus your creativity and generate more interesting and relevant rhymes.

\\ 
\\ \bottomrule
\caption{Generated Prompt Comparison for task "Rhymes"}
\label{tab:promptcompare}
 \end{longtable}

\clearpage
\section{Few-shot Add/ Removal Examples}
\label{fewshot-examples}
Below are cases listing examples where few-shot examples are added or removed with \frameNamenospace.

\subsection{Zero-shot to Few-shot}

\begin{table}[h]
\centering
\begin{center}
\begin{tabular}{ m{2cm} | m{13cm}}
\toprule
\textbf{Task} & Ethos \\ \midrule
 \textbf{Input Prompt} & Is the following text hate speech? \textbackslash n  \#\# Output Format \#\# True or False

 \\ 
 \midrule
 \textbf{Output Prompt} & Your task is to evaluate whether the provided input includes any offensive language. This could be language that is sexist, racist, discriminatory, or advocates violence or harm. Offensive language can also be identified by an aggressive or belittling tone. However, remember that the mere presence of profanity does not necessarily make a statement offensive. The context and intention behind the statement should also be taken into account. If none of these conditions are fulfilled, return 'False'. \newline

\hl{For instance, the phrase "You are a fool" is offensive due to its derogatory tone, so you should return 'True'. On the other hand, the phrase "This is a damn tasty pizza" includes profanity but is not offensive, so you should return 'False'. Similarly, a phrase like "All people of X race are lazy" is offensive because of its racist undertones, so you should return 'True'. In contrast, a phrase like "I dislike the color yellow" is not offensive, so you should return 'False'.}

\\
\bottomrule
\end{tabular}
\end{center}
\caption{Add Few-shot Example: added examples are highlighted.}
\label{tab:fewshot-additional}
\end{table}

\subsection{Zero-shot to Zero-shot}

\begin{table}[h]
\centering
\begin{center}
\begin{tabular}{ m{2cm} | m{13cm}}
\toprule
\textbf{Task} & Ethos \\ \midrule
 \textbf{Input Prompt} & Is the following text hate speech? \textbackslash n  \#\# Output Format \#\# True or False

 \\ 
 \midrule
 \textbf{Output Prompt} & Classify the given text as hate speech or not and generate a binary output of 1 for Yes and 0 for No.

\\
\bottomrule
\end{tabular}
\end{center}
\caption{Zero-shot to Zero-shot}
\label{tab:fewshot-zero}
\end{table}

\subsection{Few-shot to Zero-shot}

\begin{longtable}[c]{ m{2cm} | m{13cm}}

 \endfirsthead

 \hline
 \multicolumn{2}{l}{Continuation of Table \ref{tab:fewshot-removal}}\\
 \hline

 \endhead

\hline
 \multicolumn{2}{l}{Continuation of Table \ref{tab:fewshot-removal}}\\
 \hline

 \endfoot

 \endlastfoot

 \toprule
 \textbf{Task} & Hyperbaton \\ \midrule
 \textbf{Input Prompt} & Order adjectives correctly in English sentences.\newline
\newline

Q: Which sentence has the correct adjective order:\newline
Options: \newline
(A) rubber terrible ship\newline
(B) terrible rubber ship\newline
A: Let's think step by step.\newline
When there is more than one adjective before a noun, the adjectives need to respect the following order before a noun: "[1. opinion] [2. size] [3. age] [4. shape] [5. color] [6. origin] [7. material] [8. purpose] noun".\newline
Option (A): "rubber terrible ship". (1) rubber" falls into the material category. (2) "terrible" falls into the opinion category. Option (A) has the following adjective order: [7. material] [1. opinion] (or, in numeric terms, 7 1). Because 7 < 1 is not correct, (A) does not have the correct ordering.\newline
Option (B): "terrible rubber ship". Option (B) has the following adjective order: [1. opinion] [7. material] (or, in numeric terms, 1 7). Because 1 < 7 is correct, (B) has the correct ordering. So the answer is (B).\newline
\newline
Q: Which sentence has the correct adjective order:\newline
Options:\newline
(A) repulsive small Brazilian exercise ship\newline
(B) Brazilian repulsive exercise small ship\newline
A: Let's think step by step.\newline
When there is more than one adjective before a noun, the adjectives need to respect the following order before a noun: "[1. opinion] [2. size] [3. age] [4. shape] [5. color] [6. origin] [7. material] [8. purpose] noun".\newline
Option (A): "repulsive small Brazilian exercise ship". (1) "repulsive" falls into the opinion category. (2) "small" falls into the size category. (3) "Brazilian" falls into the origin category. (4) "exercise" falls into the purpose category. Option (A) has the following adjective order: [1. opinion] [2. size] [6. origin] [8. purpose] (or, in numeric terms, 1 2 6 8). Because 1 < 2 < 6 < 8 is correct, (A) has the correct ordering.\newline
Option (B): "Brazilian repulsive exercise small ship". Option (B) has the following adjective order: [6. origin] [1. opinion] [8. purpose] [2. size] (or, in numeric terms, 6 1 8 2). Because 6 < 1 < 8 < 2 is not correct, (B) does not have the correct ordering. So the answer is (A).\newline
...

 \\ 
 \midrule
 \textbf{Output Prompt} & Identify the sentence with the correct order of adjectives: opinion, size, age, shape, color, origin, material, purpose.

\\
\bottomrule
\caption{Few-shot to Zero-shot}
\label{tab:fewshot-removal}
 \end{longtable}

\clearpage

\begin{longtable}[c]{ m{2cm} | m{13cm}}
 \endfirsthead

 \hline
 \multicolumn{2}{l}{Continuation of Table \ref{tab:fewshot-fewshot}}\\
 \hline

 \endhead

\hline
 \multicolumn{2}{l}{Continuation of Table \ref{tab:fewshot-fewshot}}\\
 \hline

 \endfoot

 \endlastfoot

 \toprule
  \textbf{Task} & Hyperbaton \\ \midrule
 \textbf{Input Prompt} & Order adjectives correctly in English sentences.\newline
\newline

Q: Which sentence has the correct adjective order:\newline
Options: \newline
(A) rubber terrible ship\newline
(B) terrible rubber ship\newline
A: Let's think step by step.\newline
When there is more than one adjective before a noun, the adjectives need to respect the following order before a noun: "[1. opinion] [2. size] [3. age] [4. shape] [5. color] [6. origin] [7. material] [8. purpose] noun".\newline
Option (A): "rubber terrible ship". (1) rubber" falls into the material category. (2) "terrible" falls into the opinion category. Option (A) has the following adjective order: [7. material] [1. opinion] (or, in numeric terms, 7 1). Because 7 < 1 is not correct, (A) does not have the correct ordering.\newline
Option (B): "terrible rubber ship". Option (B) has the following adjective order: [1. opinion] [7. material] (or, in numeric terms, 1 7). Because 1 < 7 is correct, (B) has the correct ordering. So the answer is (B).\newline
\newline
Q: Which sentence has the correct adjective order:\newline
Options:\newline
(A) repulsive small Brazilian exercise ship\newline
(B) Brazilian repulsive exercise small ship\newline
A: Let's think step by step.\newline
When there is more than one adjective before a noun, the adjectives need to respect the following order before a noun: "[1. opinion] [2. size] [3. age] [4. shape] [5. color] [6. origin] [7. material] [8. purpose] noun".\newline
Option (A): "repulsive small Brazilian exercise ship". (1) "repulsive" falls into the opinion category. (2) "small" falls into the size category. (3) "Brazilian" falls into the origin category. (4) "exercise" falls into the purpose category. Option (A) has the following adjective order: [1. opinion] [2. size] [6. origin] [8. purpose] (or, in numeric terms, 1 2 6 8). Because 1 < 2 < 6 < 8 is correct, (A) has the correct ordering.\newline
Option (B): "Brazilian repulsive exercise small ship". Option (B) has the following adjective order: [6. origin] [1. opinion] [8. purpose] [2. size] (or, in numeric terms, 6 1 8 2). Because 6 < 1 < 8 < 2 is not correct, (B) does not have the correct ordering. So the answer is (A).\newline
...

 \\ 
 \midrule
 \textbf{Output Prompt} & Adjective Order Practice: \newline
 \newline
 In English grammar, the order of adjectives is important to convey accurate and effective descriptions. Here are two examples of sentences with adjectives. Determine which sentence has the correct adjective order. \newline
 \newline
 Example 1: \newline
 a) The big, red, round ball bounced down the street. \newline
 b) The round, red, big ball bounced down the street. \newline
 \newline
 Example 2: \newline
 a) The delicious, homemade, chocolate cake was devoured by the guests. \newline
 b) The chocolate, homemade, delicious cake was devoured by the guests. \newline
 \newline
 Choose the letter of the sentence with the correct adjective order as your answer. \newline
 \newline
 Note: Adjective order generally follows the pattern of opinion, size, age, shape, color, origin, material, and purpose. If the adjectives do not fit into this pattern, the order is determined by the speaker's preference. \newline
 \newline
 If you choose the incorrect sentence, we will explain why it is wrong to help you learn from your mistakes. Good luck!

\\
\bottomrule
\caption{Few-shot to Few-shot}
\label{tab:fewshot-fewshot}
 \end{longtable}

 \subsection{Synthetic Few-shot Examples.} 
We observe that in certain cases \frameName would generate novel synthetic few-shot examples instead of selecting from existing ones. To verify their veracity, we conduct a manual evaluation of the accuracy of the few-shot examples generated by \frameName on a total of 24 instruction deduction tasks. We find that 90 out of the 92 examples evaluated (97.8\%) are accurate. Among them, 24 out of the 92 (24.09\%) are aligned with samples present in the training set. 
There are two cases where the synthetic example is inaccurate: the sentiment of \textit{"A non-mystery mystery"} is identified as \textit{"neutral"} where the ground truth is \textit{"negative"}, and \textit{"Little more than a well-mounted history lesson"} is identified as \textit{"neutral"} where the ground truth is \textit{"negative"}. In both cases, we empirically validate that such a level of inaccuracy does not influence prompt performance (score remained $94\%$ regardless of the labels).

\clearpage

\section{Generated Prompts}
\label{sec:generated_prompt}

In this section, we list the prompts generated by \frameName with the best performance for each task. All prompts are generated by gpt-3.5. We observe a mix of few-shot prompts and zero-shot prompts for different tasks. This indicates both LLM's ability to perform in-context prompt optimization and \frameNamenospace's ability to traverse the whole problem space to find optimal solutions.

We also notice that the few-shot examples in the final prompts are largely generated by LLM instead of copied from example instruction or training sets. Thus it serves as further proof of LLM's capability of in-context prompt optimization and \frameNamenospace's credibility in this problem space.

\begin{longtable}[c]{ m{2cm} | m{12cm}}
 \endfirsthead

 \hline
 \multicolumn{2}{l}{Continuation of Table \ref{tab:bbh-prompt}}\\
 \hline

 \endhead

\hline
 \multicolumn{2}{l}{Continuation of Table \ref{tab:bbh-prompt}}\\
 \hline
 \endfoot

 \endlastfoot

 \toprule
\textbf{Causal Judgment} & Provide reactions to intentional actions in diverse scenarios, while also considering causation and its complexities. To assist with determining causation, provide specific guidelines and examples for each scenario. To avoid any confusion or misinterpretation, precise language and definitions will be used throughout the prompt. Additionally, feedback from experts and individuals with relevant experience in the field of causation will be incorporated to ensure accuracy and relevance. To challenge users' critical thinking skills, include diverse and complex scenarios that require creative problem-solving and a deeper understanding of causation in various areas of life.  \\ 
 \midrule
 \textbf{Dyke Languages} &  Correctly close all brackets, including nested brackets, in the provided sequence in the proper order from innermost to outermost. Mistakes such as forgetting to close a bracket or closing brackets in the wrong order can result in an error. If an error is made, a clear and concise message will indicate which bracket is not properly closed and suggest how to correct it. A visual representation of the correct sequence of closed brackets is provided below:  \newline
[ { ( [ { ( ) } ] ) } ] \newline
\newline
Examples of valid and invalid inputs:\newline
\newline
Valid input: [ { ( ) } ]  \newline
Valid input: [ { ( [ ] ) } ] \newline
Invalid input: [ { ( [ ) } ] \newline
Warning message: The bracket at position 8 is not properly closed. Please close the bracket to ensure proper syntax. \newline
Suggested correction: [ { ( [ ] ) } ] \newline
 \newline
Invalid input: [ { ( [ } ] ) ] \newline
Warning message: The bracket at position 8 is not properly closed.  Please close the bracket to ensure proper syntax. \newline
Suggested correction: [ { ( [ ] ) } ] \newline
 \\
    \midrule
  \textbf{Formal Fallacies}& Read the given argument carefully and determine whether it is deductively valid or invalid based on the explicitly stated premises. Provide a justification for your answer.\\
  \midrule
 \textbf{Dis-ambiguation QA} &  For each sentence with a gender-neutral pronoun, determine the antecedent or state if it is ambiguous. Use (A) for the first option, (B) for the second option, or (C) for ambiguous. Additionally, provide an explanation of the antecedent (the person or thing the pronoun refers to) for each sentence.\\
 \midrule
   \textbf{Hyperbaton}& Test your knowledge of adjective order in English sentences with interactive exercises and quizzes. Learn the rule of opinion-size-age-shape-color-origin-material-purpose noun and apply it to different types of nouns such as animals, objects, and people. Practice constructing your own sentences and receive feedback on incorrect answers to improve your skills. By the end of this exercise, you'll be able to confidently order adjectives and communicate accurately in English.\\
  \midrule
  \textbf{Logical Deduction Five}& On a plate, there are three fruits: a red apple, a yellow banana, and a green pear. The banana is positioned to the immediate left of the apple, meaning there are no other fruits between them. The pear is the rightmost fruit, meaning it comes last in the order.  \newline
\newline
Which of the following statements is true? \newline

(A) The red apple is the leftmost fruit.\newline
(B) The yellow banana is the leftmost fruit.\newline
(C) The green pear is the leftmost fruit.\newline
\newline
Explanation: \newline
To solve this prompt, pay attention to the precise language used to describe the relationships between the fruits and their positions in the order. The banana is to the immediate left of the apple, meaning it is directly adjacent to it and there are no other fruits between them. The pear is the rightmost fruit, meaning it comes last in the order. 

Therefore, the correct answer is (B) The yellow banana is the leftmost fruit. \newline
\newline
To further practice this concept, here are some additional examples: \newline
1. On a plate, there are three different colored balls: a blue ball, a red ball, and a green ball. The red ball is directly to the left of the blue ball, meaning there are no other balls between them. The green ball is the rightmost. Which of the following statements is true? \newline
(A) The blue ball is the leftmost ball.\newline
(B) The red ball is the leftmost ball.\newline
(C) The green ball is the leftmost ball.\newline
\newline
2. In a row of three houses, there is a blue house, a yellow house, and a green house. The blue house is in the middle, and the yellow house is directly to the left of the green house, meaning there are no other houses between them. Which of the following statements is true? 
(A) The blue house is the leftmost house.\newline
(B) The yellow house is the leftmost house.\newline
(C) The green house is the leftmost house.\newline
\newline
Feedback: \newline
If you selected the wrong answer, here's why: \newline
(A) The red apple is not the leftmost fruit. The yellow banana is directly to its left. \newline
(C) The green pear is not the leftmost fruit. The banana and the apple come before it in the order. \newline
\newline
To avoid confusion, use precise language to describe the relationships between objects and their positions in the order. Avoid using vague terms like "newer" or "older" without specifying their exact relationship to other objects in the order. Provide more context or details to help clarify any ambiguities in the prompt. Make sure the order of the objects is clearly defined and consistent throughout the prompt.
\\
 \midrule
 \textbf{Reasoning Colored Objects} &  Identify the color of objects arranged in a row on a surface.

Q: On the desk, there is a black stapler, a green highlighter, a yellow ruler, a blue pen, and a purple marker. What color is the pen? \newline
Options:\newline
(A) red \newline
(B) orange \newline
(C) yellow \newline
(D) green \newline
(E) blue \newline
(F) brown \newline
(G) magenta \newline
(H) fuchsia \newline
(I) mauve \newline
(J) teal \newline
(K) turquoise \newline
(L) burgundy \newline
(M) silver \newline
(N) gold \newline
(O) black \newline
(P) grey \newline
(Q) purple \newline
(R) pink \newline
A: Let's think step by step. \newline
According to this question, the objects are arranged in a row, from left to right, as follows: (1) a black stapler, (2) a green highlighter, (3) a yellow ruler, (4) a blue pen, and (5) a purple marker. \newline
The pen is the fourth item on the list, namely (4). The color of the pen is blue. So the answer is (E). \newline
\\
 \midrule

  \textbf{Salient Translation Error Detection} &  Read the following translations from German to English and identify the type of error present in each one. The error can be one of the following types: Named Entities, Numerical Values, Modifiers or Adjectives, Negation or Antonyms, Facts, or Dropped Content. Write the corresponding letter for each error type in the options provided.  \newline
 \newline
For example:  \newline
Source: Der Hund ist braun. \newline
Translation: The cat is brown. \newline
The translation contains an error pertaining to: \newline
Options: \newline
(A) Modifiers or Adjectives \newline
(B) Numerical Values \newline
(C) Negation or Antonyms \newline
(D) Named Entities \newline
(E) Dropped Content \newline
(F) Facts  \newline
 \newline
Output: (D) \newline
\\
\bottomrule
\caption{BBH Prompts}
\label{tab:bbh-prompt}
 \end{longtable}
\clearpage

 \begin{longtable}[c]{ m{2cm} | m{13cm}}
 \endfirsthead
 \hline
 \multicolumn{2}{l}{Continuation of Table \ref{tab:apo-prompt}}\\
 \hline

 \endhead
\hline
 \multicolumn{2}{l}{Continuation of Table \ref{tab:apo-prompt}}\\
 \hline

 \endfoot

 \endlastfoot

 \toprule
\textbf{Ethos} & Does the provided text contain hate speech? Return a boolean value of True or False.  \\ 
 \midrule
 \textbf{Liar} &   Analyze the context and other information provided to determine the truthfulness of the statement. To do so, consider the following guidelines: \newline
\newline
1. Identify key sources of information, such as reputable news outlets or government reports, and consider the credibility of the sources. Look for corroborating evidence and consider any potential biases or conflicts of interest. \newline
\newline
2. Conduct additional research or seek out expert opinions when necessary to determine the truthfulness of a statement. Use resources or links to relevant information provided, and consider consulting with subject matter experts or fact-checking organizations. \newline
\newline
3. Note that the determination of truthfulness may not always be possible based on the information provided, and that additional research or analysis may be required. Use your best judgment and be transparent about any uncertainties or limitations in your analysis. \newline
\newline
4. Consider specific examples or scenarios to help you apply the prompt in different contexts. For instance, you might analyze a political statement, a scientific claim, or a news article. Be aware of common pitfalls or errors, such as relying on unreliable sources or failing to consider alternative explanations.  \newline
\newline
Output Format: Assign 0 for true and 1 for false. Note that this determination is based on the information provided and may not be definitive.

 \\
    \midrule
  \textbf{Sarcasm}& Determine if the input contains any language that could be considered derogatory or discriminatory towards a particular group based on their race, ethnicity, gender, sexual orientation, religion, or any other protected characteristic. If such language is found, output True. If not, output False. The prompt should be trained on a diverse dataset to improve its accuracy and reduce errors.

\\
\bottomrule
\caption{APO Prompts}
\label{tab:apo-prompt}
 \end{longtable}

\clearpage

  \begin{longtable}[c]{ m{2cm} | m{13cm}}
 \endfirsthead

 \hline
 \multicolumn{2}{l}{Continuation of Table \ref{tab:ape-prompt}}\\
 \hline

 \endhead

 \hline
 \multicolumn{2}{l}{Continued next page for Table \ref{tab:ape-prompt}}\\
 \hline
 \endfoot

 \endlastfoot

 \toprule
\textbf{Antonyms} & "Provide a list of adjectival antonyms for each of these words, keeping in mind the given context:"\newline
\newline
\#\# Input \#\#: hot (in the context of weather)\newline
\#\# Output \#\#: ['cold', 'cool', 'chilly']\newline
\newline
\#\# Input \#\#: happy (in the context of emotions)\newline
\#\# Output \#\#: ['sad', 'unhappy', 'depressed', 'miserable']\newline
\newline
\#\# Input \#\#: big (in the context of size)\newline
\#\# Output \#\#: ['small', 'tiny', 'little', 'miniature']\newline
\newline
\#\# Input \#\#: fast (in the context of speed)\newline
\#\# Output \#\#: ['slow', 'sluggish', 'leisurely', 'gradual']\newline
\newline
\#\# Input \#\#: old (in the context of age)\newline
\#\# Output \#\#: ['young', 'new', 'fresh', 'modern'] \newline

 \\
    \midrule
  \textbf{Cause Effect}& Determine the sentence that is the cause in each pair. Remember to thoroughly comprehend the meaning of each sentence before selecting the cause. Additionally, verify your output to ensure that you only include the sentence that is the cause. To aid in identifying cause and effect relationships, consider using keywords or phrases that indicate causality, analyzing the context of each sentence, and practicing with feedback and interactive activities.
\\
\midrule
  \textbf{Common Concept} & For each input, come up with a category or characteristic that they have in common and write it as the output. Use your knowledge and experience to make educated guesses and be creative in your thinking. Also, try to keep the output concise and clear.
\\
\midrule
  \textbf{Diff} & Subtract the second number from the first number and give me the result. Make sure to double check your calculations and write the answer as a string in a list format.
\\
\midrule
  \textbf{First Word Letter} & Write a program that takes in a word and returns a list containing the first letter of the word as a string. The program will be used to label items in a game. \newline
\newline
Make sure to handle cases where the input word is empty or only contains whitespace. You can use the string method `strip()` to remove any leading or trailing whitespace. If the input is empty or contains only whitespace, return an empty list.\newline
\newline
To ensure that your program works correctly, test it with the following examples:\newline
\newline
Example 1:\newline
Input: "apple"\newline
Output: ["a"]\newline
\newline
Example 2:\newline
Input: "  banana"\newline
Output: ["b"]\newline
\newline
Example 3:\newline
Input: ""\newline
Output: []\newline
\newline
Example 4:\newline
Input: "  "\newline
Output: []\newline
\\
\midrule
  \textbf{Informal Formal} & Reword the following sentences using more formal language, but also provide alternative rewordings that are more appropriate for different contexts:\newline
\newline
1. "Regrettably, I am unable to attend the meeting tomorrow." (formal) \newline
Alternative: "Unfortunately, I won't be able to make it to the meeting tomorrow." (casual)\newline
\newline
2. "I must depart now, farewell!" (overly formal) \newline
Alternative: "I have to go now, see you later!" (casual)\newline
\newline
3. "I apologize, but I am unable to assist you with that matter." (formal) \newline
Alternative: "I'm sorry, but I can't help you with that." (casual)\newline
\newline
4. "Thank you for the invitation, however, I am unable to attend." (formal) \newline
Alternative: "Thanks for inviting me, but I can't make it." (casual)\newline
\newline
5. "In my opinion, this is the optimal choice." (formal) \newline
Alternative: "I think this is the best option." (casual)\newline

\\
\midrule
  \textbf{Large Animal} & Choose one animal as the output based on its size. For example, if the input pair is "elephant, mouse", choose "elephant" as the output. If the input pair is "giraffe, lion", choose "giraffe" as the output. Use the following criteria to choose the output:\newline 
\newline
- If one animal is significantly larger than the other, choose the larger animal as the output. \newline
- If the animals are similar in size, choose the animal with the name that comes first alphabetically as the output. \newline
\newline
Here are some examples of correct outputs: \newline
\newline
- "whale, dolphin" -> choose "whale" as the output \newline
- "panda, koala" -> choose "panda" as the output \newline
- "tiger, zebra" -> choose "tiger" as the output \newline
\newline
Choose the output carefully to avoid confusion and errors.

\\
\midrule
  \textbf{Letters List} & Please write a program that takes in a word as input and outputs a list of its letters separated by spaces. The output should be a list with one element containing the separated letters in the same order as the input word.\newline
\newline
To ensure the program works correctly, please follow these guidelines:\newline
\newline
1. Input validation: Check that the input is a non-empty string containing only alphabetic characters. If the input is invalid, print an error message and exit the program.\newline
\newline
2. Separating the letters: Use the `split()` method to separate the letters of the input word. \newline
\newline
3. Expected output format: The output should be a list with one element containing the separated letters in the same order as the input word.\newline
\newline
Here are some examples of valid and invalid input:\newline
\newline
Valid input: "hello"\newline
Expected output: ["h", "e", "l", "l", "o"]\newline
\newline
Invalid input: "hello world"\newline
Expected output: "Error: Input must be a non-empty string containing only alphabetic characters."\newline
\newline
Invalid input: "123"\newline
Expected output: "Error: Input must be a non-empty string containing only alphabetic characters."

\\
\midrule
  \textbf{Taxonomy Animal} & "List all the animals from the given inputs." \newline
\newline
\#\# Input \#\#: apple, banana, orange, kiwi, grape \newline
\#\# Output \#\#: []\newline
\newline
\#\# Input \#\#: dog, cat, fish, bird, hamster \newline
\#\# Output \#\#: ['dog', 'cat', 'fish', 'bird', 'hamster']\newline
\newline
\#\# Input \#\#: elephant, giraffe, lion, tiger, zebra \newline
\#\# Output \#\#: ['elephant', 'giraffe', 'lion', 'tiger', 'zebra']\newline
\newline
\#\# Input \#\#: pencil, eraser, notebook, ruler, pen \newline
\#\# Output \#\#: []\newline
\newline
\#\# Input \#\#: turtle, snake, lizard, frog, salamander \newline
\#\# Output \#\#: ['turtle', 'snake', 'lizard', 'frog', 'salamander']

\\
\midrule
  \textbf{Negation} & For each input, negate the specified part of the statement and write it as an output. \newline
\newline
1. Negate the part about using the gold color: "We will use gold as the primary color for our new logo." Output: "We will not use gold as the primary color for our new logo."\newline
\newline
2. Negate the part about Gary Kubiak participating as a player: "Gary Kubiak will play as a quarterback in the upcoming game." Output: "Gary Kubiak will not play as a quarterback in the upcoming game."\newline
\newline
Note: When negating statements with proper nouns or names, simply negate the verb or action associated with the noun or name.

\\
\midrule
  \textbf{Num Verbal} & Convert a given number into its English word representation, including commas for thousands and negative sign if applicable.\newline
\newline
\#\# Input 1 \#\# : 1234 \newline
\#\# Output 1 \#\#: ['one thousand two hundred and thirty-four']\newline
\newline
\#\# Input 2 \#\# : 987654321 \newline
\#\# Output 2 \#\#: ['nine hundred and eighty-seven million six hundred and fifty-four thousand three hundred and twenty-one']\newline
\newline
\#\# Input 3 \#\# : 0 \newline
\#\# Output 3 \#\#: ['zero']\newline
\newline
\#\# Input 4 \#\# : -42 \newline
\#\# Output 4 \#\#: ['negative forty-two']\newline
\newline
\#\# Input 5 \#\#: 999999999 \newline
\#\# Output 5 \#\#: ['nine hundred and ninety-nine million nine hundred and ninety-nine thousand nine hundred and ninety-nine']

\\
\midrule
  \textbf{Active Passive} & Passive Voice Practice:
\newline
In passive voice, the subject of the sentence receives the action instead of performing it. Rewrite each sentence in passive voice.\newline
\newline
Example: The dog chased the cat.\newline
Passive voice: The cat was chased by the dog\newline
\newline
1. The teacher graded the exams.\newline
2. The company launched a new product.\newline
3. The chef cooked a delicious meal.\newline
4. The team won the championship.\newline
5. The doctor prescribed medication for the patient.\newline
\newline
Instructions:\newline
- Rewrite each sentence in passive voice.\newline
- Make sure the subject of the sentence receives the action instead of performing it.\newline
- Use the examples provided to guide you.\newline
- Check your work for accuracy and clarity.\newline
\newline
Feedback:\newline
- If you have any questions or need clarification, please ask.\newline
- Practice makes perfect! Keep practicing to improve your writing skills.\newline
- If you make any mistakes, don't worry! Learn from them and try again

\\
\midrule
  \textbf{Singular Plural} &Add an "s" or the correct plural form to the end of the input word, depending on the following rules:\newline
\newline
1. If the word ends in "y" with a consonant before it, change the "y" to "ies" instead of just adding an "s".\newline
2. If the word ends in "f" or "fe", change the "f" or "fe" to "ves" instead of just adding an "s".\newline
3. If the word is already plural, return the input word as is instead of adding an "s".\newline
4. If the word has an irregular plural form, return the correct plural form instead of just adding an "s".\newline
\newline
Examples:\newline
\newline
- Input: cat\newline
  Output: cats\newline
\newline
- Input: book\newline
  Output: books\newline
\newline
- Input: car\newline
  Output: cars\newline
\newline
- Input: tree\newline
  Output: trees\newline
\newline
- Input: computer\newline
  Output: computers\newline
\newline
- Input: story\newline
  Output: stories\newline
\newline
- Input: half\newline
  Output: halves\newline
\newline
- Input: aircraft\newline
  Output: aircraft\newline
\newline
- Input: century\newline
  Output: centuries

\\
\midrule
  \textbf{Rhymes} & Create a list of words that rhyme with the given word. To ensure that your rhymes are accurate, make sure that the words have the same vowel sound and ending consonant sound. For example, "cat" rhymes with "bat" and "hat," but not with "dog" or "mat." \newline
\newline
To get started, here are some examples of words that rhyme with the given word: \newline
\newline
- Love: dove, glove, above, shove, of \newline
- Time: rhyme, chime, climb, mime, prime \newline
\newline
To find more rhyming words, you can use a rhyming dictionary, online resources, or brainstorm with friends. Be creative and try to use a variety of different rhyming words instead of repeating the same one multiple times. \newline
\newline
To avoid common pitfalls, make sure to double-check your spelling and pronunciation of the words. Also, avoid using words that only partially rhyme or have a different stress pattern. \newline
\newline
After you've created your list, ask for feedback on the quality of your rhymes. This can help you to improve and refine your skills. \newline
\newline
For an added challenge, consider generating rhyming words that fit a particular theme or context. This can help you to focus your creativity and generate more interesting and relevant rhymes.

\\
\midrule
  \textbf{Second Word Letter} & For each input word with at least two letters, identify and output the second letter. Please ensure that the input is a valid word in the specified language or dialect to prevent errors. The prompt is case-insensitive, so it will work for both uppercase and lowercase letters. \newline
\newline
Examples: \newline
- Input: "hello" Output: "e"\newline
- Input: "apple" Output: "p"\newline
- Input: "book" Output: "o"\newline
\newline
Please note that the language or dialect of the input should be specified to avoid confusion with words that have different spellings or pronunciations in different regions.

\\
\midrule
  \textbf{Sentence Similarity} & Rate the similarity of two given sentences on a scale of 1 to 5, where 1 indicates a significant difference in meaning and 5 indicates almost identical meaning. Please consider the following factors when rating: \newline
\newline
- The overall message and purpose of the sentences \newline
- The structure and syntax of the sentences \newline
- The use of key words and phrases \newline
\newline
Provide a brief explanation for your rating, taking into account any minor differences in wording or details that may affect the similarity rating. Additionally, please provide context for the sentences being compared, such as the intended audience or purpose. \newline
\newline
For reference, here are some examples of sentences that fall into each category: \newline
\newline
Highly similar: "The cat sat on the mat" and "The mat was sat on by the cat" \newline
Moderately similar: "I enjoy playing soccer" and "Soccer is a fun sport to play" \newline
Not similar at all: "The sky is blue" and "I am going to the beach tomorrow" \newline
\newline
Thank you for your evaluation and explanation.

\\
\midrule
  \textbf{Sentiment} & Please analyze the following statements and determine their overall sentiment as either ['negative', 'neutral', 'positive']. Keep in mind the context and any figurative language used.\newline
\newline
1. The sun is shining and the birds are singing.\newline
Output: ['positive']\newline
\newline
2. I failed my exam and now I have to retake the class.\newline
Output: ['negative']\newline
\newline
3. My best friend surprised me with a thoughtful gift.\newline
Output: ['positive']\newline
\newline
4. The traffic on the highway was backed up for miles.\newline
Output: ['negative']\newline
\newline
5. I received a promotion at work and a raise in salary.\newline
Output: ['positive']\newline
\newline
6. A non-mystery mystery.\newline
Output: ['neutral']\newline
\newline
7. Little more than a well-mounted history lesson.\newline
Output: ['neutral']\newline
\newline
8. Too daft by half ... but supremely good natured.\newline
Output: ['positive']\newline 
\newline
Note: This prompt uses more sophisticated language analysis techniques to better understand the sentiment of the input. However, providing more context for the input is still important for accurate sentiment analysis.

\\
\midrule
  \textbf{Orthography Starts With} & SIdentify the first word or phrase that starts with the letter given in the input. The identified word or phrase should not contain any punctuation or special characters, and should be case-insensitive. If there are no words or phrases starting with the given letter, return an empty list.\newline
\newline
Here are the input-output pairs:\newline
\newline
Input: She sang a beautiful song to the audience. [b]\newline
Output: ['beautiful']\newline
\newline
Input: The cat chased the mouse. [c]\newline
Output: ['cat']\newline
\newline
Input: It is important to always be kind to others. [i]\newline
Output: ['important']\newline
\newline
Input: The dog barked loudly, frightening the neighbors. [l]\newline
Output: ['loudly']\newline
\newline
Input: The book is on the shelf. [s]\newline
Output: ['shelf']\newline
\newline
Input: The baby cried all night. [n]\newline
Output: []\newline
\newline
Input: The teacher gave a long lecture on the history of art. [l]\newline
Output: ['lecture']\newline
\newline
Input: The car drove down the street, passing by many shops. [s]\newline
Output: ['street']\newline
\newline
Input: To the boy's delight, he received a new toy for his birthday. [t]\newline
Output: ['toy']\newline
\newline
Note: If there are multiple words or phrases starting with the given letter, the prompt will return the first one encountered. If the input contains multiple sentences or clauses, the prompt will identify the first word or phrase that starts with the given letter in the entire input text. The output will be in lowercase

\\
\midrule
  \textbf{Sum} & "Write a program that takes two numbers as input and returns their sum as a string in a list. Make sure to test your program with different inputs to ensure it works correctly. Remember to convert the input numbers to integers before adding them together, and then convert the sum back to a string before putting it in a list. Also, make sure to use the correct syntax for creating a list with one element (i.e. use square brackets around the string). Good luck!"

\\
\midrule
  \textbf{Synonym} & Please provide a list of synonyms for the given words that convey a similar meaning and are commonly used in everyday language. Be sure to double-check your spelling and grammar before submitting. \newline
\newline
For example, if the word is "happy," acceptable synonyms could be "joyful," "pleased," or "content."\newline
\newline
Please use gender-neutral language and avoid using words with different connotations or meanings. If you notice any incorrect synonyms, please flag them and provide feedback for improvement.\newline
\newline
Words to avoid using as synonyms include those with different connotations or meanings, such as "ecstatic" for "happy" or "depressed" for "sad."

\\
\midrule
  \textbf{Trans En De} & Translate the following English words into German.\newline
\newline
\#\# Input \#\# : happy \newline
\#\# Output \#\#: ['glücklich']\newline
\#\# Input \#\# : love \newline
\#\# Output \#\#: ['Liebe']\newline
\#\# Input \#\# : cat \newline
\#\# Output \#\#: ['Katze']\newline
\#\# Input \#\# : dog \newline
\#\# Output \#\#: ['Hund']\newline
\#\# Input \#\# : house \newline
\#\# Output \#\#: ['Haus']\newline
\#\# Input \#\# : tree \newline
\#\# Output \#\#: ['Baum']\newline
\#\# Input \#\# : water \newline
\#\# Output \#\#: ['Wasser']\newline
\#\# Input \#\# : sun 
\#\# Output \#\#: ['Sonne']\newline
\#\# Input \#\# : moon \newline
\#\# Output \#\#: ['Mond']\newline
\#\# Input \#\# : star \newline
\#\# Output \#\#: ['Stern']

\\
\midrule
  \textbf{Trans En Es} & Convert these English terms into their corresponding Spanish translations.\newline 
\newline
\#\# Input \#\# : happy \newline
\#\# Output \#\#: ['feliz'] \newline
\#\# Input \#\# : beach \newline
\#\# Output \#\#: ['playa'] \newline
\#\# Input \#\# : computer \newline
\#\# Output \#\#: ['computadora'] \newline
\#\# Input \#\# : book \newline
\#\# Output \#\#: ['libro'] \newline
\#\# Input \#\# : music \newline
\#\# Output \#\#: ['música']

\\
\midrule
  \textbf{Trans En Fr} & Translate the following English words into French. \newline
\newline
\#\# Input \#\# : happy\newline
\#\# Output \#\#: ['heureux'] \newline
\#\# Input \#\# : love\newline
\#\# Output \#\#: ['amour'] \newline
\#\# Input \#\# : family\newline
\#\# Output \#\#: ['famille'] \newline
\#\# Input \#\# : friend \newline
\#\# Output \#\#: ['ami'] \newline
\#\# Input \#\# : music\newline
\#\# Output \#\#: ['musique'] \newline
\#\# Input \#\# : beach \newline
\#\# Output \#\#: ['plage'] \newline
\#\# Input \#\# : book \newline
\#\# Output \#\#: ['livre'] \newline
\#\# Input \#\# : movie \newline
\#\# Output \#\#: ['film'] \newline
\#\# Input \#\# : food\newline
\#\# Output \#\#: ['nourriture'] \newline
\#\# Input \#\# : travel \newline
\#\# Output \#\#: ['voyage']

\\
\midrule
  \textbf{Word In Context} & Compare the usage of a given word in two different sentences and determine if they have the same or different meanings based on the context of the sentences. Write "same" or "not the same" as the output. \newline
\newline
To avoid ambiguity and ensure clarity, please provide sufficient context for the sentences. If the word has multiple meanings depending on the context, please indicate all correct answers. \newline
\newline
For example, consider the word "bank." In the sentence "I need to deposit my paycheck at the bank," and "I sat on the bank of the river and watched the sunset," the word "bank" has different meanings. Therefore, the correct answer would be "not the same." \newline
\newline
Please note that the comparison should be based on the context of the sentences, not just the isolated word
\\
\bottomrule
\caption{APE Prompts}
\label{tab:ape-prompt}
 \end{longtable}

\clearpage

\end{document}